\documentclass{article}

\PassOptionsToPackage{numbers,sort&compress}{natbib}
\usepackage[final]{neurips_2024}

\usepackage{algpseudocode}
\usepackage{mathtools}
\usepackage{algorithm}
\usepackage{graphicx}
\usepackage{booktabs}
\usepackage{amsmath}
\usepackage{amsbsy}
\usepackage{soul}
\usepackage{enumitem}
\usepackage{amssymb}
\usepackage{graphicx}
\usepackage{booktabs}
\usepackage{multirow}
\usepackage{siunitx} 
\usepackage{diagbox}
\usepackage{fancyhdr}
\usepackage{subcaption}
\usepackage{gensymb}
\usepackage{wrapfig}
\usepackage{xr-hyper}
\usepackage{minitoc}
\usepackage{setspace}
\usepackage{xspace}
\newcolumntype{P}[1]{>{\centering\arraybackslash}p{#1}}
\newcolumntype{M}[1]{>{\centering\arraybackslash}m{#1}}

\newcommand{\argmin}{\mathop{\mathrm{argmin}}}

\newcommand{\norm}[1]{\left\lVert#1\right\rVert}

\newcommand{\mypara}[1]{\noindent\textbf{#1.}~}

\usepackage[utf8]{inputenc} 
\usepackage[T1]{fontenc}    
\usepackage{hyperref}     

\usepackage{url}            
\usepackage{booktabs}       
\usepackage{amsfonts}       
\usepackage{nicefrac}      
\usepackage{microtype}     
\usepackage{xcolor}

\title{PrefPaint: Aligning Image Inpainting Diffusion Model with Human Preference}

\author{
    Kendong Liu\textsuperscript{1}\thanks{Equal Contribution} , \quad
    Zhiyu Zhu\textsuperscript{1}$^{*}$\thanks{Corresponding Author.} , \quad
    Chuanhao Li\textsuperscript{2}$^{*}$ , \quad 
    Hui Liu\textsuperscript{3} , \quad \\
    \textbf{Huanqiang Zeng\textsuperscript{4} ,}  \quad
    \textbf{Junhui Hou\textsuperscript{1}} \quad \\
    \textsuperscript{1}City University of Hong Kong~
    \textsuperscript{2}Yale University~
    \textsuperscript{3}Saint Francis University~
    \textsuperscript{4}Huaqiao University\\
    \texttt{\{kdliu2-c, zhiyuzhu2-c\}@my.cityu.edu.hk}\\
\texttt{chuanhao.li.cl2637@yale.edu}\quad\texttt{h2liu@sfu.edu.hk}\quad\texttt{jh.hou@cityu.edu.hk} \\
}

\begin{document}
\doparttoc
\faketableofcontents

\maketitle

\begin{abstract}
In this paper, we make the first attempt to align diffusion models for image inpainting with human aesthetic standards via a reinforcement learning framework, significantly improving the quality and visual appeal of inpainted images. Specifically, instead of directly measuring the divergence with paired images, we train a reward model with the dataset we construct, consisting of nearly 51,000 images annotated with human preferences. Then, we adopt a reinforcement learning process to fine-tune the distribution of a pre-trained diffusion model for image inpainting in the direction of higher reward. Moreover, we theoretically deduce the upper bound on the error of the reward model, which illustrates the potential confidence of reward estimation throughout the reinforcement alignment process, thereby facilitating accurate regularization.
Extensive experiments on inpainting comparison and downstream tasks, such as image extension and 3D reconstruction, demonstrate the effectiveness of our approach, showing significant improvements in the alignment of inpainted images with human preference compared with state-of-the-art methods. This research not only advances the field of image inpainting but also provides a framework for incorporating human preference into the iterative refinement of generative models based on modeling reward accuracy, with broad implications for the design of visually driven AI applications. Our code and dataset are publicly available at \url{https://prefpaint.github.io}.
\end{abstract}

\section{Introduction}
\begin{figure}[t]
    \centering
    \resizebox{1.0\textwidth}{!}{\includegraphics{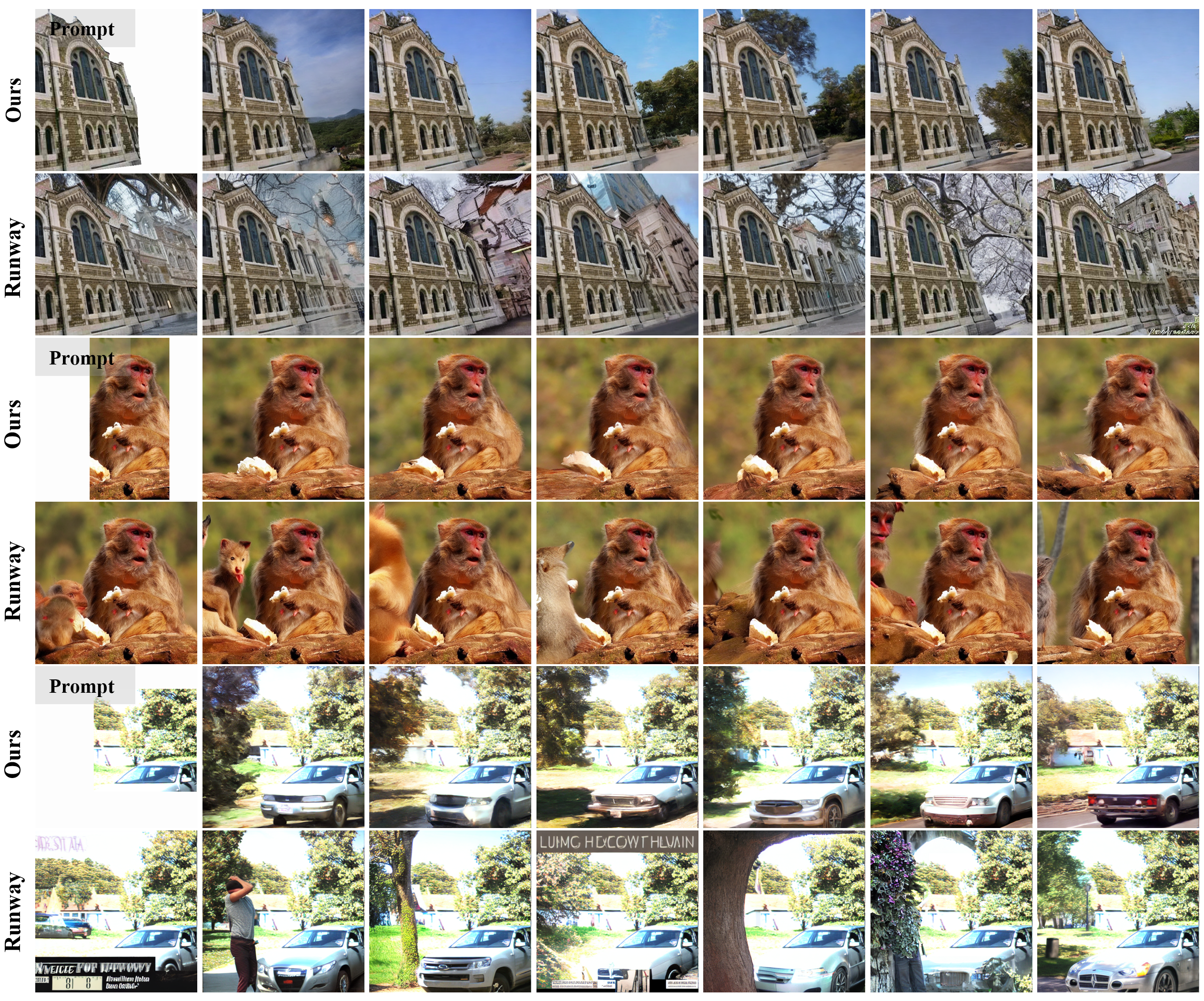}} \vspace{-0.5cm}
    \caption{Visual comparisons of the results by the diffusion-based image inpainting model named ``\textit{Runway}", and the aligned model through the proposed method.}
    \vspace{-4mm}
    \label{fig:InpaintingSample}
\end{figure}

Image inpainting, the process of filling in missing or damaged parts of images, is a critical task in computer vision with applications ranging from photo restoration~\cite{yi2020contextual,li2020recurrent} to content creation~\cite{anciukevivcius2023renderdiffusion,svitov2023dinar}. Traditional approaches have leveraged various techniques, from simple interpolation~\cite{karaca2016interpolation,arias2012nonlocal,alsalamah2016medical} to complex texture synthesis~\cite{guo2021image,jain2023keys}, to achieve visually plausible results.  The recent advent of deep learning, particularly diffusion models, has revolutionized the field by enabling more coherent and contextually appropriate inpaintings~\cite{lahiri2020prior,zhang2022gan,anciukevivcius2023renderdiffusion,yang2023uni}. Despite these advancements, a significant gap remains between the technical success of these models and their alignment with human aesthetic preferences, which are inherently subjective and intricate. As shown in Fig.~\ref{fig:InpaintingSample}, the existing stable diffusion-based inpainting model tends to generate weird and discord reconstruction.

Human preference for visual content is influenced by complex and mutual factors, including but not limited to personal background, experiences, and the context in which the content is viewed~\cite{wu2023human,satgunam2013factors}. This makes the task of aligning inpainting models with human preference particularly challenging, as it requires the model not only to understand the content of the missing parts but also to predict and adapt to the diverse tastes of its users. 

Inspired by recent advancements in the reinforced alignment of large pre-trained models~\cite{shinn2024reflexion,rafailov2024direct,stiennon2020learning,casper2023open,munos2023nash}, we propose to align diffusion models for image inpainting with human preference through reinforcement learning. Specifically, our approach is grounded in the hypothesis that by incorporating human feedback into the training loop, we can guide the model toward generating inpainted images that are not only technically proficient but also visually appealing to users. Technically, we formulate the boundary of each reward prediction in terms of the model's alignment with human aesthetic preferences, thereby leveraging the accuracy of the reward model to amplify the regularization strength on more reliable samples. Extensive experiments validate that the proposed method can consistently reconstruct visually pleasing inpainting results and greatly surpass state-of-the-art methods.

In summary, the main contributions of this paper lie in: \vspace{-0.2cm}

\begin{itemize}[itemsep=1pt, topsep=0pt, leftmargin=15pt]
    \item  we make the \textit{first attempt} to align diffusion models for image inpainting with human preferences by integrating human feedback through reinforcement learning;
    \vspace{-1mm}
    \item we theoretically deduce the accuracy bound of the reward model, modulating the refinement process of the diffusion model for robustly improving both efficacy and efficiency; and
    \vspace{-1mm}
    \item we construct a dataset containing 51,000 inpainted images annotated with human preferences.
  
\end{itemize}

\vspace{-1mm}
\section{Related Work}
\mypara{Reinforcement Learning \& Model Alignment}
Reinforcement learning~\cite{kaelbling1996reinforcement,arulkumaran2017deep,franccois2018introduction} is a paradigm of machine learning where an agent learns to make decisions by taking actions in an environment to maximize some notion of cumulative reward. The foundational theory of reinforcement learning is rooted in the concepts of Markov decision processes \cite{feinberg2012handbook}, which provide a mathematical structure for modeling decision-making in environments with stochastic dynamics and rewards~\cite{van2012reinforcement,gattami2021reinforcement}. With the advent of deep learning, deep reinforcement learning~\cite{arulkumaran2017deep,wang2020deep} has significantly expanded the capabilities and applications of traditional reinforcement learning. The integration of neural networks with reinforcement learning, exemplified by the Deep Q-Network~\cite{ong2015distributed,mnih2013playing} algorithm, has enabled the handling of high-dimensional state spaces, which were previously intractable. Subsequent innovations, including policy gradient methods~\cite{lu2021decentralized,khadka2018evolution} like Proximal Policy Optimization~\cite{schulman2017proximal} and actor-critic frameworks~\cite{bahdanau2016actor,fujimoto2018addressing} like Soft Actor-Critic~\cite{haarnoja2018soft}, have further enhanced the efficiency and stability of learning in complex environments.

The recent surge in popularity of large-scale models has significantly underscored the importance of reinforcement learning in contemporary AI research and application~\cite{shinn2024reflexion,rafailov2024direct}. As these models, including large language models~\cite{kasneci2023chatgpt,liu2023gpt} and deep generative networks~\cite{ruthotto2021introduction,ho2020denoising,sohl2015deep}, become more prevalent, reinforcement learning is increasingly employed to fine-tune, control, and optimize their behaviors in complex, dynamic environments. This integration is particularly visible in areas such as natural language processing, where reinforcement learning techniques are used to improve the conversational abilities of chat-bots and virtual assistants, making them more adaptive and responsive to user needs.~\cite{stiennon2020learning,casper2023open,munos2023nash} Moreover, in the realm of content recommendation and personalization, reinforcement learning algorithms are instrumental in managing the balance between exploration of new content and exploitation of known user preferences, significantly enhancing the user experience. The growing intersection between large models and reinforcement learning not only pushes the boundaries of what's achievable in AI but also amplifies the need for sophisticated reinforcement learning techniques that can operate at scale, adapt in real-time, and make decisions under uncertainty, thereby marking a pivotal evolution in how intelligent systems are developed and deployed.

\mypara{Diffusion Model}
Diffusion models have recently emerged as a powerful class of generative models~\cite{ho2020denoising,sohl2015deep,song2020denoising,saharia2022palette,ho2022imagen}, demonstrating remarkable success in generating high-quality, coherent images~\cite{hou2024global,you2024nvs}. These models work by gradually transforming a distribution of random noise into a distribution of images, effectively 'diffusing' the noise into structured patterns~\cite{song2023consistency,song2020score}. In the context of image inpainting, diffusion models offer a significant advantage by leveraging their generative capabilities to predict and fill in missing parts of images in a way that is contextually and visually coherent with the surrounding image content~\cite{lugmayr2022repaint,corneanu2024latentpaint}.

Recent studies have showcased the potential of diffusion models in achieving state-of-the-art results in image inpainting tasks, outperforming previous generative models like Generative Adversarial Networks in terms of image quality and coherence~\cite{stypulkowski2024diffused,muller2022diffusion}. However, while these models excel in technical performance, there remains a gap in their ability to cater to diverse human aesthetic preferences. Most existing works focus on the objective quality of inpainting results, such as fidelity to the original image and coherence of the generated content, with less attention given to subjective satisfaction or preference alignment.

\vspace{-2mm}
\section{Proposed Method}
\vspace{-2mm}

Due to the random masking within the task of image inpainting, there may not be a definitive causal relationship between the known and inpainted contents, which manifests as a one-to-many issue. Consequently, stringent per-pixel regularization inherently results in unnatural reconstruction, which significantly diverges from samples conforming to human preference.
To this end, we propose a reinforcement learning-based alignment process involving human preferences to fine-tune pre-trained diffusion models for image inpainting, aiming to improve the visual quality of inpainted images (Sec.~\ref{sec:reinforce}). 

More importantly, we theoretically deduce the upper bound on the error of the reward model (Sec.~\ref{sec:bounding}). Based on this deduction, we formulate a reward trustiness-aware alignment process that is more efficient and effective (Sec.~\ref{sec:RewardTrust}).

\subsection{Reinforced Training of Diffusion Models for Image Inpainting}
\label{sec:reinforce}
Diffusion models~\cite{ho2020denoising,song2020denoising} iteratively refine a randomly sampled standard Gaussian distributed noise, resulting in a generated image. To adjust the distribution of diffusion models, we introduce human feedback rewards to measure and regularize the distribution of model sampling outputs. To be specific, instead of applying standard policy-gradient descent~\cite{kakade2001natural} that is hard to converge to high-quality models, inspired by classical methods, e.g., TRPO/PPO~\cite{schulman2015trust,schulman2017proximal}, which introduces a model trust region to avoid potential model collapse during the training process, we achieve the reinforced training of a diffusion model as
\begin{equation}
    \nabla_{\boldsymbol{\theta} }\mathcal{J}(\boldsymbol{x}) = - \int_{P_{\boldsymbol{\theta}'}} \frac{\nabla_{\boldsymbol{\theta}} P_{\boldsymbol{\theta}}(\boldsymbol{x})}{P_{\boldsymbol{\theta}'}(\boldsymbol{x})} \mathcal{R}(\boldsymbol{x}) + \kappa \nabla_{\boldsymbol{\theta}} \mathcal{D}(P_{\boldsymbol{\theta}'}|P_{\boldsymbol{\theta}}),
\end{equation}
where $\mathcal{R}(\cdot)$ represents the reward model, which quantifies the quality of diffusion samples; $\boldsymbol{x}$ is the reconstructed sample; $P_{\boldsymbol{\theta}}$ and $P_{\boldsymbol{\theta}'}$ represent the probabilistic functions of training and reference models parameterized with $\boldsymbol{\theta}$ and $\boldsymbol{\theta}'$, respectively; $\nabla_{\boldsymbol{\theta}}$ calculates the derivative on $\boldsymbol{\theta}$; $\mathcal{D}(\cdot|\cdot)$ is the divergence measurement of the given two distributions for regularizing the distribution shifting; the hyperparameter $\kappa$ balances the two terms. Inspired by recent work~\cite{zhang2024large,xu2024imagereward}, we take the same step to measure the probability of each diffusion step via calculating the probability of perturbation noise, e.g., the DDIM sampling algorithm~\cite{song2020denoising}, 
\begin{equation}
    \boldsymbol{x}_{t-1} = \underbrace{\sqrt{\alpha_{t-1}}\left(\frac{\boldsymbol{x}_t-\sqrt{1-\alpha_t}\epsilon^{(t)}_{\theta}(\boldsymbol{x}_t)}{\sqrt{\alpha_t}}\right) + \sqrt{1-\alpha_{t-1}-\sigma_t^2}\epsilon^{(t)}_{\theta}(\boldsymbol{x}_t)}_{\Bar{\boldsymbol{x}}_{t-1}} + \underbrace{\sigma_t\epsilon_t \vphantom{\frac{\boldsymbol{x}_t-\sqrt{1-\alpha_t}\epsilon^{(t)}_{\theta}(\boldsymbol{x}_t)}{\sqrt{\alpha_t}}}}_{\widetilde{\boldsymbol{x}}_{t-1} },
\end{equation}
where $\epsilon^{(t)}_{\boldsymbol{\theta}}(\cdot)$ is the noise estimate diffusion network; $t$ is the diffusion step; the scalar $\alpha_t$ controls the noise to signal ratio; and $\sigma_t\epsilon_t$ represents a Gaussian noise to increase the sampling diversity. Thus, the result of each reverse step consists of a deterministic component $\Bar{x}_{t-1}$ and a probabilistic component $\widetilde{\boldsymbol{x}}_{t-1}$. Since $\boldsymbol{x}_{t-1}\sim \mathcal{N}(\Bar{\boldsymbol{x}}_{t-1},\sigma_t \mathbf{I})$, we calculate the density of each step $P = \Phi\left(\frac{\boldsymbol{x}_{t-1}-\Bar{\boldsymbol{x}}_{t-1}}{\sigma_t}\right)$ to approximate the probability, where $\Phi(\cdot)$ denotes the density function of standard Gaussian distribution.

\subsection{Bounding Reward Model Error}
\label{sec:bounding}

\begin{wrapfigure}[18]{r}{0.4\textwidth}
    \vspace{-0.5cm}
    \centering
    \includegraphics[width=0.4\textwidth]{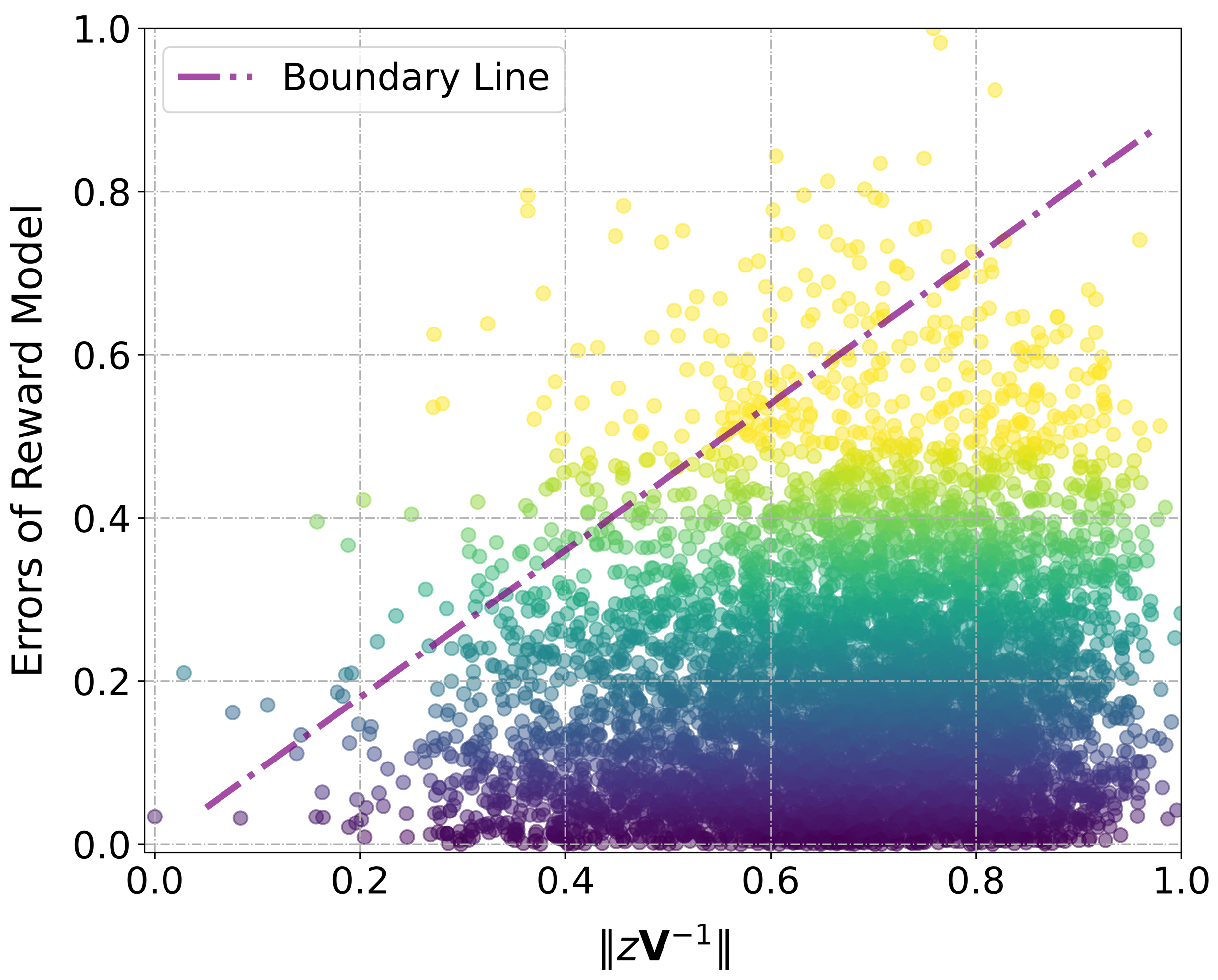}
    \caption{\label{Fig_Bound}Experimental plot of reward prediction error \textit{vs.} $\| z\|_{\mathbf{V}^{-1}}$ on the validation set, where a dashed line is an upper boundary of error, positively relative to $\| z\|_{\mathbf{V}^{-1}}$.}
\end{wrapfigure}

The precision of the reward model $\mathcal{R}(\cdot)$ assumes a pivotal function within this learning framework as it directs the optimization trajectory. In this section, we theoretically derive its error upper bound, which can facilitate the reinforced training process in terms of both efficacy and efficiency.

Denote by $\mathbf{X}=[\boldsymbol{x}_1, \boldsymbol{x}_2, \cdots, \boldsymbol{x}_N]\in\mathbb{R}^{N\times D}$ and $\boldsymbol{y}=[y_1, y_2,~\cdots,~y_N]\in\mathbb{R}^{N\times 1}$ a set of $N$ inpainted images of dimension $D$ by diffusion models and corresponding ground-truth reward values from human experts, respectively.
By dividing the reward model into two parts, i.e., feature extractor $\mathcal{F}(\cdot)$ and linear regression weights $\boldsymbol{\psi} \in \mathbb{R}^{D'\times 1}$, we thus can represent it as $\left< \mathcal{F}(\boldsymbol{x}_t),\boldsymbol{\psi} \right>$, where $\left< \cdot,\cdot \right>$ calculates the inner product of two vectors. 
We can formulate the learning process of the last weights as the result of the following optimization process:
\begin{equation}
    \hat{\boldsymbol{\psi}}  = \argmin\limits_{\boldsymbol{\psi}}\norm{ \bigl \langle \mathcal{F}(\mathbf{X}),\boldsymbol{\psi} \bigr \rangle - \boldsymbol{y} }_2^2 + \lambda \| \boldsymbol{\psi} \|_2^2,
\end{equation}
\noindent where $\|\cdot\|_2$ is the $\ell_2$ norm of a vector, $\lambda>0$, and the second term is used for alleviating the over-fitting phenomenon. Thus, we have
\begin{align}
    \hat{\boldsymbol{\psi}}=& (\mathbf{Z}^\textsf{T}\mathbf{Z} + \lambda \mathbf{I})^{-1}\mathbf{Z}^\textsf{T}\boldsymbol{y},\\
    \hat{\boldsymbol{\psi}} - \boldsymbol{\psi}_* =&(\mathbf{Z}^\textsf{T}\mathbf{Z} + \lambda \mathbf{I})^{-1} \mathbf{Z}^\textsf{T} \boldsymbol{\zeta} - \lambda (\mathbf{Z}^\textsf{T}\mathbf{Z} + \lambda \mathbf{I})^{-1} \boldsymbol{\psi}_*,
\end{align}
where $\mathbf{Z}\in \mathbb{R}^{N \times D'} = \mathcal{F}(\mathbf{X})$ is the set of embeddings of $\mathbf{X}$. $\boldsymbol{\psi}_*$ is the ideal weight with $\boldsymbol{y} = \mathbf{Z} \boldsymbol{\psi}_* + \boldsymbol{\zeta}$ and $\boldsymbol{\zeta}$ a noise term between $\mathbf{Z} \boldsymbol{\boldsymbol{\psi}}_*$ and $\boldsymbol{y}$. Then, the following term also holds:
\begin{align}
    \boldsymbol{z}^\textsf{T}\hat{\boldsymbol{\psi}} - \boldsymbol{z}^\textsf{T}\boldsymbol{\psi}_* =&\boldsymbol{z}^\textsf{T}(\mathbf{Z}^\textsf{T}\mathbf{Z} + \lambda \mathbf{I})^{-1} \mathbf{Z}^\textsf{T} \boldsymbol{\zeta} - \lambda \boldsymbol{z}^\textsf{T} (\mathbf{Z}^\textsf{T}\mathbf{Z} + \lambda \mathbf{I})^{-1} \boldsymbol{\psi}_*,
\end{align}
where $\boldsymbol{z}\in\mathbb{R}^{D' \times 1}= \mathcal{F}(\boldsymbol{x})$ is the reward embedding of a typical input sample.
Based on Cauchy-Schwarz inequality, there is 
\begin{align}
    |\boldsymbol{z}^\textsf{T}\hat{\boldsymbol{\psi}} - \boldsymbol{z}^\textsf{T}\boldsymbol{\psi}_* | \leq \| \boldsymbol{z}\|_{\mathbf{V}^{-1}}(\|\mathbf{Z}^\textsf{T} \zeta \|_{\mathbf{V}^{-1}} + \lambda^{1/2}\|\boldsymbol{\psi}_* \|_2),
\end{align}
where $\mathbf{V} = \mathbf{Z}^\textsf{T}\mathbf{Z} + \lambda \mathbf{I}$ and $\|\boldsymbol{z}\|_{\mathbf{V}^{-1}} := \|\boldsymbol{z}^\textsf{T}\mathbf{V}^{-1}\|_2$.
Moreover, based on \textit{Theorem 2} of \cite{abbasi2011improved}, for $\delta > 0$, we have $1-\delta$ probability to make following inequality stands 
\begin{align}
    \| \mathbf{Z}^\textsf{T} \epsilon \|_{\mathbf{V}^{-1}} \leq B \sqrt{2 \log(\frac{\texttt{det}(\mathbf{V})^{1/2}\texttt{det}(\lambda \mathbf{I})^{-1/2}}{\delta})},
\end{align}
\noindent where $\texttt{det}(\cdot)$ computes the determinant of the matrix, and $B$ is a scalar. We have 
\begin{align}
    \label{eq:Error}
    | \boldsymbol{z}^\textsf{T}\hat{\boldsymbol{\psi}} - \boldsymbol{z}^\textsf{T}\boldsymbol{\psi}_* | \leq \| \boldsymbol{z}\|_{\mathbf{V}^{-1}}\underbrace{\left(B \sqrt{2 \log(\frac{\texttt{det}(\mathbf{V})^{1/2}\texttt{det}(\lambda \mathbf{I})^{-1/2}}{\delta})} + \lambda^{1/2} \|\boldsymbol{\psi}_* \|_2\right)}_{C_{bound}}.
\end{align}
Due to $C_{bound}$ being constant after the training of the reward model, we can conclude that
\begin{equation}
\label{equ:bound}
 \sup_{z\sim p(z)} | \boldsymbol{z}^\textsf{T}\hat{\boldsymbol{\psi}} - \boldsymbol{z}^\textsf{T}\boldsymbol{\psi}_* | \propto \| \boldsymbol{z}\|_{\mathbf{V}^{-1}},
\end{equation}
where $\sup$ represents the upper bound. Such a theoretical bound is also experimentally verified in Fig. \ref{Fig_Bound}.

\begin{figure}[t]
    \centering
    \resizebox{1.0\textwidth}{!}{\includegraphics{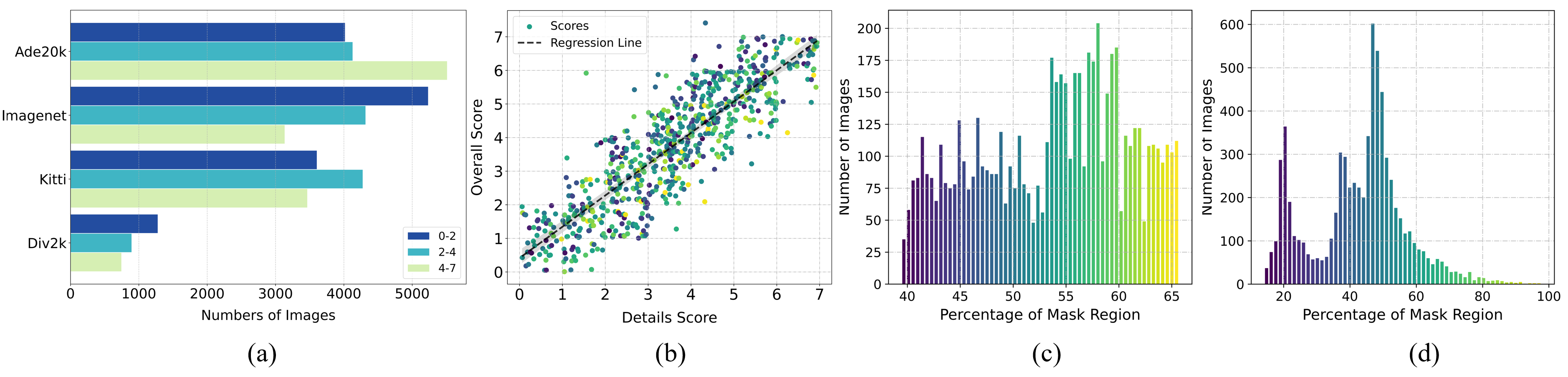}}
    \vspace{-0.5cm}
    \caption{Statistical characteristics of the dataset we constructed. \textbf{(a)}  the score distribution of the images across different selected datasets; \textbf{(b)}  the comparison between the distribution of the average score and score for details;  \textbf{(c)} and \textbf{(d)} show the numbers of images with different mask ratios on the outpainting and warping splits, respectively.}
    \label{fig:DatasetStatistic}
\end{figure}

\subsection{Reward Trustiness-Aware Alignment Process}
\label{sec:RewardTrust}
According to Eq. \eqref{equ:bound}, it can be known that the error of the reward model is bounded by $\| \boldsymbol{z}\|_{\mathbf{V}^{-1}}$, there might be a relatively large error for those $\| \boldsymbol{z} \|_{\mathbf{V}^{-1}}$ quite large. Thus, we further propose a weighted regularization strategy to amplify the penalty strength of those samples in the high-trust region. Specifically, we calculate the amplification factor as
\begin{equation}
    \label{Eq:Scaling}
    \gamma = e^{- k\| \boldsymbol{z}\|_{\mathbf{V}^{-1}} + b},
\end{equation}
where the hyperparameters $k=0.05$ and $b=0.7$ are used for scaling the regularization strength. 
We then define the overall gradient of the reward trustiness-aware alignment process as 
\begin{equation}
    \label{eq:out}
    \nabla_{\boldsymbol{\theta}} \mathcal{J}'(\boldsymbol{x}) = \gamma \nabla_{\boldsymbol{\theta}} \mathcal{J}(\boldsymbol{x}) ,
\end{equation}
where $\mathcal{J}'(\mathbf{\boldsymbol{x}})$ is the weighted reward loss for sample $\mathbf{x}$. For the updating of the reference model $\boldsymbol{\theta}'$, we adopt~\cite{zhang2024large} to update the model in each optimization step.  Correctly amplifying the scaling factor can both speed up the convergence speed and reconstruction effectiveness, as experimentally validated in Sec.~\ref{Sec:Ablation}.

\section{Human Preference-Centric Dataset for Reward} 
\label{Sec:Reward}

We first randomly selected 6,000, 4,000, 6,000, and 1,000 images with diverse content from ADE20K~\cite{zhou2017scene,zhou2019semantic}, ImageNet~\cite{deng2009imagenet}, KITTI~\cite{Geiger2012CVPR}, and Div2K~\cite{Agustsson_2017_CVPR_Workshops,Timofte_2018_CVPR_Workshops} datasets, respectively. We then applied the operations described below to generate prompt images (i.e., incomplete images), which were further fed into the diffusion model for image inpainting named \textit{Runway}~\cite{runway}, producing inpainted images. To mitigate the potential bias of the reward model on different rewards, we repetitively generated from a given prompt image three distinct inpainted images. Consequently, we obtained 51,000 inpainted images in total, which were scored by human experts following the criteria described below.

\begin{table*}[t]
\renewcommand{\arraystretch}{0.95}
\setlength{\tabcolsep}{5.7pt} 
\centering
\small
\caption{\label{Tab:comparisons} Quantitative comparisons of different methods. $\star$ indicates the small model~(non SD-based); ``$S$'' is the number of sampling times. For the calculation of WinRate, we first derive the best sample of the compared method among $S$ sampling times. Then, we calculate it as $\frac{T_w}{T}$, where $T_w$ indicates the number of compared samples that surpass the results of \textit{Runway} ($S=1$) and $T$ is the total number of prompts.  ``$\uparrow$ (resp. $\downarrow$)" means the larger (resp. smaller), the better. We normalized the predicted reward values with the dataset distribution.   
``Var'' calculates the variance of different sampling times, showing the consistency of generation quality. (See the \textit{Supplementary Material} for more details.)} \vspace{-0.2cm}
\begin{tabular}{lcccM{0.68cm}c|cccM{0.68cm}c} 
\toprule
\multirow{1}{*}{Prompt Methods} & \multicolumn{5}{c|}{Outpainting Prompts} & \multicolumn{5}{c}{Warping Prompts} \\
 \cmidrule{2-11} 
\multirow{2}{*}{Metrics}  & \multicolumn{3}{c}{WinRate (\%)~$\uparrow$} & \multicolumn{2}{c|}{Reward}& \multicolumn{3}{c}{WinRate (\%)~$\uparrow$} & \multicolumn{2}{c}{Reward} \\ 
& $S$ = 1 & $S$ = 3  & $S$ = 10 & Mean$\uparrow$ & Var$\downarrow$ & $S$ = 1  & $S$ = 3  & $S$ = 10 &Mean$\uparrow$ & Var$\downarrow$  \\ \midrule
 Runway~\cite{runway} & -- -- &  73.40 & 89.32 & -- -- &0.07 & -- -- & 75.74   & 91.42 & -- -- &0.06  \\ \midrule
SD v1.5~\cite{sd1.5} & 11.95 & 20.67   & 30.24  & -0.43&0.05 & 11.38& 21.22  &32.85 &-0.38 &0.06  \\
SD v2.1~\cite{sd2.1}  &10.73 & 18.51 & 26.82 & -0.44 &0.04  & 11.68& 22.11 & 34.22 & -0.36 & 0.06 \\
SD xl~\cite{sdxl} &14.56  &22.58  &31.09 &-0.31  &0.04  &15.43 &25.43 &36.77  &-0.26  &0.05  \\ 
SD xl ++~\cite{sdxlInpt} & 21.15 & 33.25   &45.51 &-0.13 &0.05 &18.66  &30.53   &43.07  &-0.18 &0.04  \\
Compvis~\cite{compvis}  & 50.51 & 66.39  & 78.21 & +0.03 &0.03 &47.35 &65.08   &78.01 & -0.01 & 0.04 \\
Kandinsky~\cite{kandinsky}  & 14.06 &22.73   &32.16 &-0.37 & 0.04 &11.38 & 19.46  & 29.20 &-0.42 &0.05  \\ 
MAT $\star$~\cite{MAT}  &15.06  &17.97   &20.51 & -0.40 &0.01  &7.17 &9.97  &12.96 &-0.56 &0.01   \\ 
Palette $\star$~\cite{saharia2022palette}   &10.96  &16.92  &21.37 & -0.38 &0.02 &13.41  &20.18 &27.37 &-0.34  &0.03 \\ \midrule
\textbf{{Ours}} & \textbf{70.16} & \textbf{84.65}  & \textbf{93.14} & \textbf{+0.38} & \textbf{0.01} & \textbf{72.38} & \textbf{87.10}  & \textbf{93.85} &\textbf{+0.36} & \textbf{0.01} \\\bottomrule
\end{tabular}
\end{table*}

\begin{figure}[]
   \centering
  \begin{minipage}[b]{0.47\textwidth}
    \centering
    \renewcommand{\arraystretch}{1.4}
    \setlength{\tabcolsep}{1.4pt} 
    \small
    \begin{tabular}{l|ccM{0.6cm}M{0.7cm}cc|c}
    \toprule
    \multirow{2}{*}{{\diagbox[dir=NW,width=4em, height=3em, innerleftsep=0pt,innerrightsep=0pt]{Model}{Metric}}}
    &T2I & CLIP  & BLIP  & Aes.  & CA  & IS & Rank \\ 
    & ~\cite{xu2024imagereward}  
    & ~\cite{clip}  & ~\cite{blip}  & ~\cite{schuhmann2022laion} & ~\cite{szegedy2016rethinking}  & ~\cite{salimans2017pixelcnn++} &  \\\midrule
    SDv1.5 & -1.67 & 0.19  & 0.44 & 4.52 & 0.38 & 17.07 & 5.17 \\
    SDv2.1 &-1.37 & 0.20 &0.45 &4.62 &0.39 &17.07 & 4.33\\
    Kand. &-3.49 & 0.18 &0.39 &\textbf{5.19} &0.39 &17.06 & 5.33\\
    SD xl ++ & 0.63 & 0.21  & 0.46 & 4.77 & 0.40 & 18.95 & 3.17\\
    Runway & 3.16 & 0.22  & 0.48 & 4.61 & 0.43 & 20.30 & 2.33 \\
    Platte & -1.76 & 0.22  & 0.46 & 4.08 & 0.37 & 16.24 & 5.33\\
    \textbf{Ours}  & \textbf{4.49} & \textbf{0.23}  & \textbf{0.49} & 4.55 & \textbf{0.45} & \textbf{23.71} & \textbf{1.67}\\ \bottomrule
    \end{tabular}
    \captionof{table}{\label{Tab_fid} Comparison across metrics: higher values are better for all metrics except "Rank".}
    \vspace{-0.33cm}
  \end{minipage}
  \hfill
  \begin{minipage}[b]{0.50\textwidth}
    \centering
    \includegraphics[width=\textwidth]{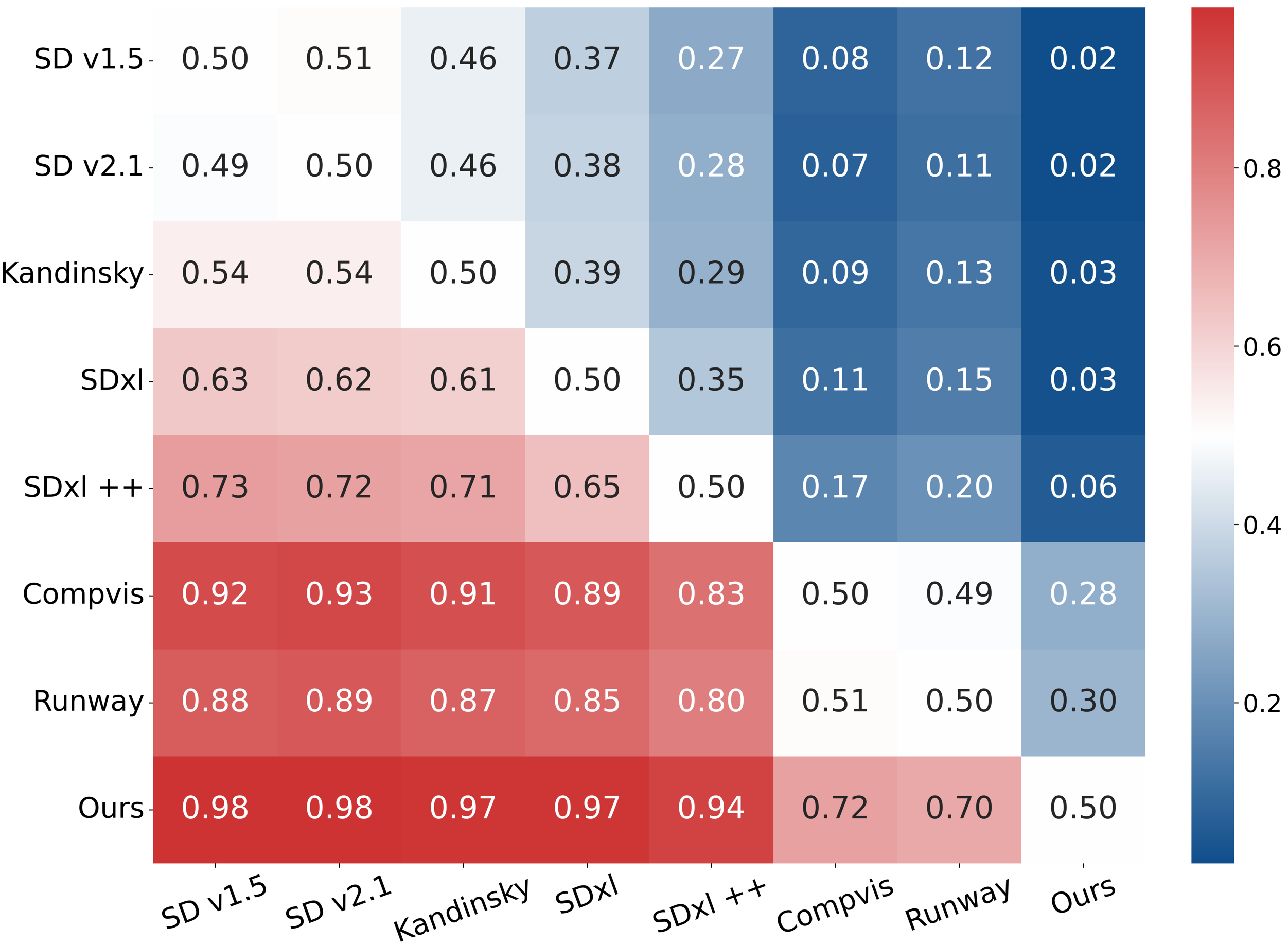}
    \vspace{-0.55cm}
    \caption{\label{Fig:WinRate}WinRate comparison heat-map between different methods.}   
    \vspace{-0.33cm}
  \end{minipage}
\vspace{-0.2cm}
\end{figure}

\noindent \textbf{Generation of Incomplete Images.} We considered two distinctive image completion patterns: inpainting and outpainting. For inpainting, we simulated warping holes on images by changing the viewpoints, where we derived the depth of the scene using \cite{yang2024depth}.
Following past practice~\cite{xiang20233d}, we defined a camera sequence that forms a sampling grid with three columns for yaw and three rows for pitch. The resulting nine views feature yaw angles within a ±0.3 range (i.e., a total range of 35\degree) and pitch angles within a ±0.15 range (i.e., a total range of 17\degree). As the range of views increases, the task of inpainting becomes progressively more challenging.
For outpainting, we randomly masked the boundary of the image through two types of random cropping methods: 
(1) square cropping, where the size of the prompt ranges from 15\% to 25\% of the image size ($512 \times 512$) randomly; (2) rectangular cropping, where the height of the prompt matches the image size, while the width is randomly sampled between 35\% and 40\% of the image size. Each cropping method accounts for half of the outpainting prompts.

\begin{figure}[t]
    \centering
    \resizebox{1.0\textwidth}{!}{\includegraphics{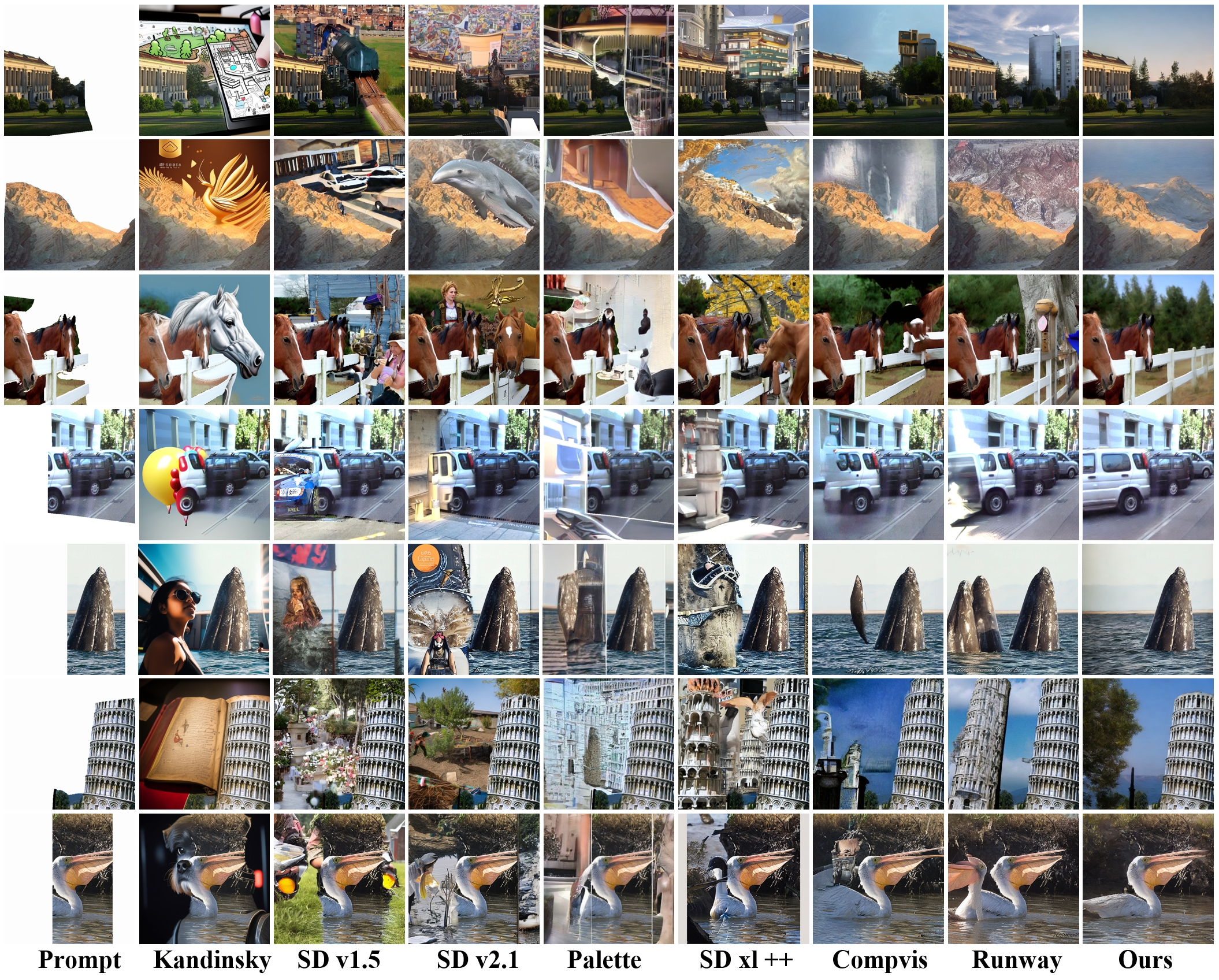}}
     \vspace{-0.4cm}
    \caption{Visual comparisons of our approach and SOTA methods. The prompted images of $5^{th}$ and $7^{th}$ rows are generated by boundary cropping, while the remaining rows by warping. All images were generated with the same random seeds. }
    \vspace{-0.8cm}
    \label{fig:Comparison}
\end{figure}

\noindent \textbf{Scoring Criteria.} In the scoring process, we employed three criteria, i.e., (\textbf{1}) \textit{structural rationality} representing the rationality of the overall structure, whether illogical objects and structures are generated; (\textbf{2}) \textit{feeling of local texture} showing whether strange textures are generated that do not conform to the characteristics of the object; and (\textbf{3}) \textit{overall feeling} indicating the impression capturing the overall feeling upon first glances at the image.  The score value is in the range of 1 to 7, indicating from the least to most favorable. The final score for reward model training is derived by averaging those three scores using the weights of $[0.15, 0.15, 0.7]$. We refer readers to the \textit{Supplementary Material} for more details about dataset process and labeling scheme.

\noindent \textbf{Statistical Characteristics of our Dataset.} As illustrated in Fig.~\ref{fig:DatasetStatistic}~(\textcolor{red}{a}), the score distribution of images from different datasets is generally uniform. Meanwhile, from Fig.~\ref{fig:DatasetStatistic}~(\textcolor{red}{b}), we can see that the details and overall score are independent of each other, showing the correctness of the scoring scheme and the necessity of each score. Finally, Figs.~\ref{fig:DatasetStatistic}~(\textcolor{red}{c}) and (\textcolor{red}{d}) show the ratio of images with different mask sizes, where it can be seen that the outpainting has a more uniform distribution than warping based hole, which rely on depth map and may not have uniform distribution.

\begin{figure}[t]
  \centering
\captionof{table}{\label{Tab_boundary} \textbf{Left}: Ablation studies on amplification factors, where "static" refers to employing a constant factor to replace $\gamma$ in Eq. \eqref{eq:out}. The column "Factor" indicates an average magnitude of amplification strength, i.e., $\mathbb{E}_{\mathbf{z}\sim \mathbf{p}(\mathbf{z})}(\gamma)$. For our method, we coordinate the value of $k$ in Eq. \eqref{Eq:Scaling} to change the $\mathbb{E}(\gamma)$ shown in the Table. "Acl." signifies acceleration, calculated by $\frac{T_{b}}{T_{m}} - 1$ with $T_{b}$ and $T_{m}$ being the convergence iterations of baseline and compared methods, respectively. For all metrics, the larger, the better. \textbf{Right}: Performance of the reward model trained with two manners based on pre-trained CLIP~\cite{clip} with various fix rates (FRs).  ``Acc'' and ``Var'' stand for  the accuracy and variance of the reward estimation, respectively. ``Bd.'' is the ratio of data below the same upper boundary. (The \underline{underlined settings} are selected.)
}
  \hspace{-2mm}
  \begin{minipage}[b]{0.48\textwidth}
    \centering
    \renewcommand{\arraystretch}{1.3}
    \setlength{\tabcolsep}{3.5pt} 
    \small
    \vspace{-2mm}
    \resizebox{1\textwidth}{!}{
    \begin{tabular}{cccc|cccc}
    \toprule
    & RL &\shortstack{Amp.} &Factor & WinRate & Reward  & \textbf{Acl.} \\  \midrule
    \textbf{a)} & $\color{gray} \times$  &\textcolor{gray}{- -} &\textcolor{gray}{- -} & \textcolor{gray}{50.00\%} & \textcolor{gray}{0.00}  & \textcolor{gray}{- -} \\ \midrule
   \textbf{b)} & $\checkmark$ & $\times$ &1.0 & 69.57\% & +0.36 & 00.00\%\\ 
    \textbf{c)}& $\checkmark$ & Static &1.4 & 65.95\% & +0.34 & +84.28\%  
    \\
    \textbf{d)}& $\checkmark$ & Static &1.6 & 65.11\% & +0.35 & 
    \textbf{+203.53} \%\\ 
    \textbf{e)}& $\checkmark$ & Ours &1.4 & \underline{\textbf{71.27\%}} & \underline{\textbf{+0.37}} & \underline{+106.40\%} \\ 
    \textbf{f)}& $\checkmark$ & Ours &1.6 & 70.47\% & +0.36 & +102.83\% \\ 
    \bottomrule
\end{tabular}}
  \end{minipage}
  \hfill
  \begin{minipage}[b]{0.51\textwidth}
    \centering
    \renewcommand{\arraystretch}{1.35}
    \setlength{\tabcolsep}{1.25pt} 
    \footnotesize
    \vspace{-0.15cm}
    \resizebox{1\textwidth}{!}{
    \begin{tabular}{lccccc|ccc|ccc} 
    \toprule
    \multicolumn{1}{l}{Training} &  & \multicolumn{4}{c|}{Classification-driven} & \multicolumn{6}{c}{Regression-driven} \\ \cmidrule{3-12} 
    \multicolumn{1}{l}{Width} &  & \multicolumn{2}{c|}{MLP-128} & \multicolumn{2}{c|}{MLP-256} & \multicolumn{3}{c|}{MLP-128} & \multicolumn{3}{c}{MLP-256} \\ \cmidrule{3-12} 
    \multicolumn{1}{l}{Metrics $\uparrow$} &  & Acc. & Var.  & Acc. & Var.  & Acc. & Var.  & \multicolumn{1}{c|}{Bd.} & Acc. & Var. & \multicolumn{1}{c}{Bd.} \\ \midrule
    \multicolumn{1}{l}{FR. = 0.3}&   & 73.49 & 0.24  & 74.29 & 0.18  & 74.68 & 0.41 & 92.1 & 74.88 & 0.38 & 93.2 \\
    \multicolumn{1}{l}{FR = 0.5}  &  & 74.73 & 0.25  & \textbf{75.32} & \textbf{0.26}  & 74.78 & 0.44 & \textbf{92.3} & 75.32 & 0.44 & 94.4 \\ 
    \multicolumn{1}{l}{FR = 0.7}  & & \textbf{74.89} & \textbf{0.28}  & 74.53 & 0.25  & \textbf{75.91} & \textbf{0.46}  & 85.0 &\underline{\textbf{75.94}} & \underline{\textbf{0.45}} & \underline{97.3}  \\
    \multicolumn{1}{l}{FR = 0.9}   &  & 74.09 & 0.16  & 72.76 & 0.14  & 74.73 & 0.41 & 78.8 & 74.83 & 0.40 & \textbf{98.0}\\ \bottomrule
    \end{tabular}}
  \end{minipage}
  \vspace{-0.4cm}
\end{figure}

\vspace{-0.3cm}
\section{Experiments}
\vspace{-0.3cm}
\noindent \textbf{Evaluation Metrics.} We adopted seven metrics to measure the quality of inpainted images, i.e., the predicted reward value by our trained reward model 
, T2I reward~\cite{xu2024imagereward}, CLIP~\cite{clip}, BLIP~\cite{blip}, Aesthetic (Aes.)~\cite{schuhmann2022laion}, Classification Accuracy (CA)~\cite{szegedy2016rethinking}, and Inception Score (IS)~\cite{salimans2017pixelcnn++}. Specifically, T2I reward, CLIP, BLIP, and Aes. directly measure the consistency between the semantics of inpainted images and the language summary of corresponding prompt images. While, CA and IS indicate the quality of the generative model. Due to the fact that our inpainting reward directly measures the reconstruction quality, we adopted it as the principle measurement of our experiment.

\noindent \textbf{Implementation Details.} We partitioned the dataset in Sec.~\ref{Sec:Reward} into training, validation and testing sets, containing 12,000, 3,000 and 2,000 prompts (with 36,000, 9,000 and 6,000 images), respectively. 
The reinforcement fine-tuning dataset contains the prompts from the original reward training dataset. We employed the pre-trained CLIP (ViT-B) checkpoint as the backbone of our reward model $\mathcal{R}(\cdot)$ with the final MLP channel equal to 256. 
We utilized a cosine schedule to adjust the learning rate. Notably, we achieved optimal preference accuracy by fixing 70\% of the layers with a learning rate of $1e-5$ and a batch size of 5. We trained the reward model with four 
\begin{wraptable}[12]{r}{0.47\textwidth}
      \centering
    \renewcommand{\arraystretch}{1.4}
    \setlength{\tabcolsep}{0.75pt} 
    \scriptsize
    \vspace{-0.5mm}
    \caption{\label{Tab_exp_functions} Performance under various parameterizations of amplification factor $\gamma$, where  $f=\| \boldsymbol{z}\|_{\mathbf{V}^{-1}}$. (The underlined settings are selected.)}
   \vspace{-1.5mm}
   \begin{tabular}{c|c|c|c|c}
    \toprule
    Amplification factor ($\gamma$)  & k & b & WinRate$\uparrow$ & Reward$\uparrow$  \\
    \midrule
     \underline{$e^{- kf + b}$}  & 0.050  & 0.70 & \textbf{71.27\%} & \textbf{0.37}  \\
    $e^{- kf + b}$  & 0.065 & 0.90 &  70.47\% & 0.36 \\
    $e^{- kf} + b_1/f + b_2$  & 0.100 & \{0.10, 0.85\} &70.07\% & 0.37\\
    $e^{- kf} + b_1/f + b_2$  & 0.120 & \{0.80, 0.85\} &69.95\%  & 0.36\\
    $-kf + b$   &1.900 &0.06 & 60.28\%& 0.28  \\
    $b$ & --  &1.43 & 65.95\% &  0.34\\
    \bottomrule
\end{tabular}
\end{wraptable}
NVIDIA GeForce RTX 3090 GPUs, each equipped with 20GB of memory. With the trained reward model, we subsequently fine-tuned the latest diffusion-based image inpainting model, namely \textit{Runway}~\cite{runway}, on four 40GB NVIDIA GeForce RTX A6000 GPUs as our PrefPaint. During fine-tuning, we employed half-precision computations with a learning rate of $2e-6$, and a batch size of 16.

\vspace{-0.25cm}
\subsection{Results of Image Inpainting} 
\vspace{-0.25cm}
\label{Sec:completion}
We compared our PrefPaint with SOTA methods both quantitatively and qualitatively to demonstrate the advantage of our method. 
We compared the WinRate and Reward under various sampling steps in Table \ref{Tab:comparisons} and Fig.~\ref{Fig:WinRate}, where it can be seen that our method actually greatly improves the probability of high-quality human-preferred samples, since with a single inference step, our method achieves more than a $70\%$ WinRate, which is similar to the baseline model (\textit{Runway}) with $S=3$. 
The higher reward value and lower variance also indicate that our method can consistently generate high-quality samples. Comparisons with other metrics in Table~\ref{Tab_fid} further support the superiority of the proposed method.
In addition, we visually compared the inpainted images by different methods in Fig.~\ref{fig:Comparison}, where it can be seen that our method can generate more meaningful and reasonable content, which is consistent with the style of the prompt region.  We refer readers to the \textit{Supplementary Material} for more details about more comparison results. 

\begin{figure}[t]
    \centering
    \resizebox{1.0\textwidth}{!}{\includegraphics{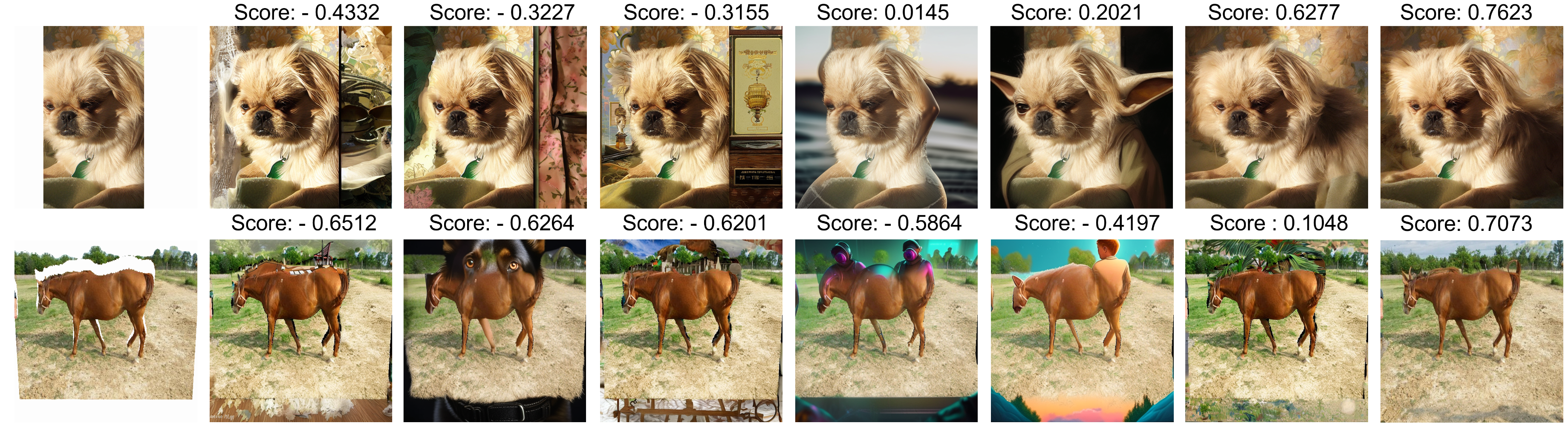}}
     \vspace{-0.3cm}
    \caption{Visualization of various image in-painting results and associated rewards from our model. Our model effectively evaluates in-painting reconstructions based on human preference. }
    \vspace{-0.2cm}
    \label{fig:reward_visualization}
\end{figure}

Finally, we have also carried out a user study to evaluate our superiority. We have randomly selected about 130 groups of results and conducted a user study involving 10 users. The WinRate map, as depicted in Fig.~\ref{Fig:WinRate_usrstudy}, demonstrates that our reward scoring is highly accurate and capable of assessing the results of inpainting under criteria based on human preference. We also present some examples scored by our reward model in Fig.~\ref{fig:reward_visualization} to illustrate the scoring criteria system.

\subsection{Ablation Study}
\label{Sec:Ablation}

\noindent \textbf{Reward Trustiness-Aware Scheme.} We validated the advantages of our reward trustiness-aware training scheme. As shown on the left side of  Table~\ref{Tab_boundary}~\textbf{b)} and \textbf{e)}, we can see that this scheme improves the reward by 1.7\% and accelerates the algorithm efficiency of 106.40\%. Although a constant amplification, such as \textbf{c)} and \textbf{d)} in Table~\ref{Tab_boundary}, can expedite the training process even faster than the proposed method, it compromises model accuracy, as evidenced by the reduced WinRate and Reward. This demonstrates the necessity and efficacy of adaptively managing the sampled trajectory and underscores the superiority of the proposed method.

\noindent \textbf{Training Manner of Reward Model.} We investigated two types of strategies to train the reward model $\mathcal{R}(\cdot)$ i.e., classification-driven and regression-driven. 
Specifically, the former classifies the discrete scores of annotated reward samples, while the latter directly makes a regression on the scores. 
As listed on the right side of Table~\ref{Tab_boundary}, it can be seen that the regression-driven training generally outperforms the classification-driven one on reward accuracy. Moreover, the larger variances from the regression-driven training show the strong discriminative ability. 
Based on accuracy and boundary performance, we finally selected a fixed rate of 0.7 and MLP-256 with the regression-driven training manner as the configuration of our reward model.

\noindent \textbf{Parameterize amplification factor $\gamma$.} To parameterize the amplification factor in Eq.~\eqref{equ:bound}, we investigated various functions to parameterize as shown in Table~\ref{Tab_exp_functions}. The experimental results indicate that the exponential function provides the best regularization effect. In contrast, the linear function and static constant do not fully exploit the regularization effect of the reward upper boundary.

\begin{figure}[t]
    \centering
    \resizebox{\textwidth}{!}{\includegraphics{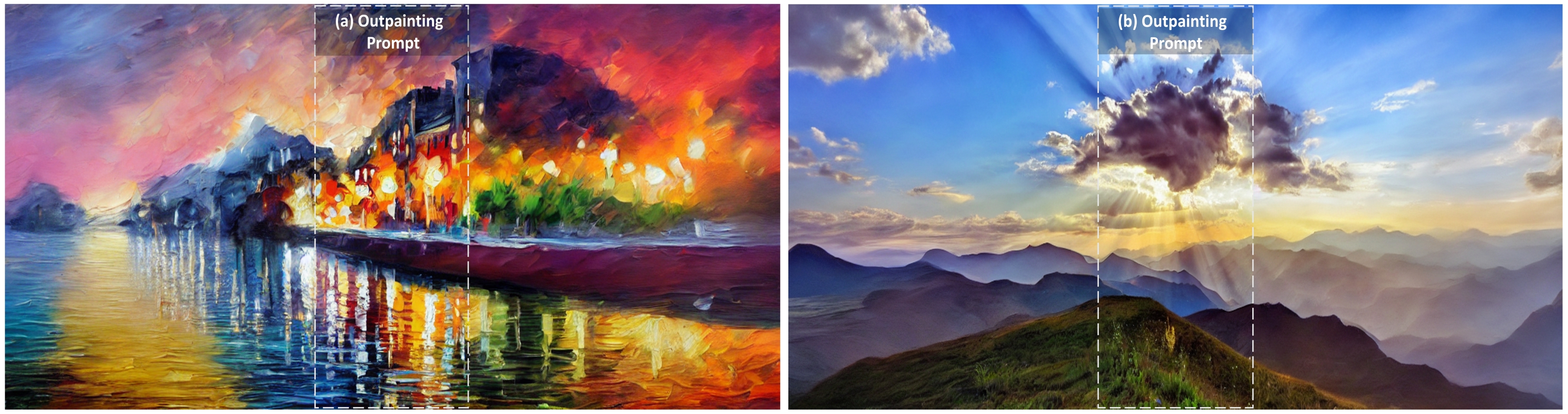}}
    \caption{Results of image FOV enlargement by our method on two scenes (\textbf{a}) and (\textbf{b}), where the prompt region (the given image) is delineated by the central area between white dashed lines.} 
    \label{fig:exp-FOV}
    \vspace{-0.1cm}
\end{figure}

\begin{figure}[t]
    \centering
    \includegraphics[height=6.2cm, width=\linewidth]{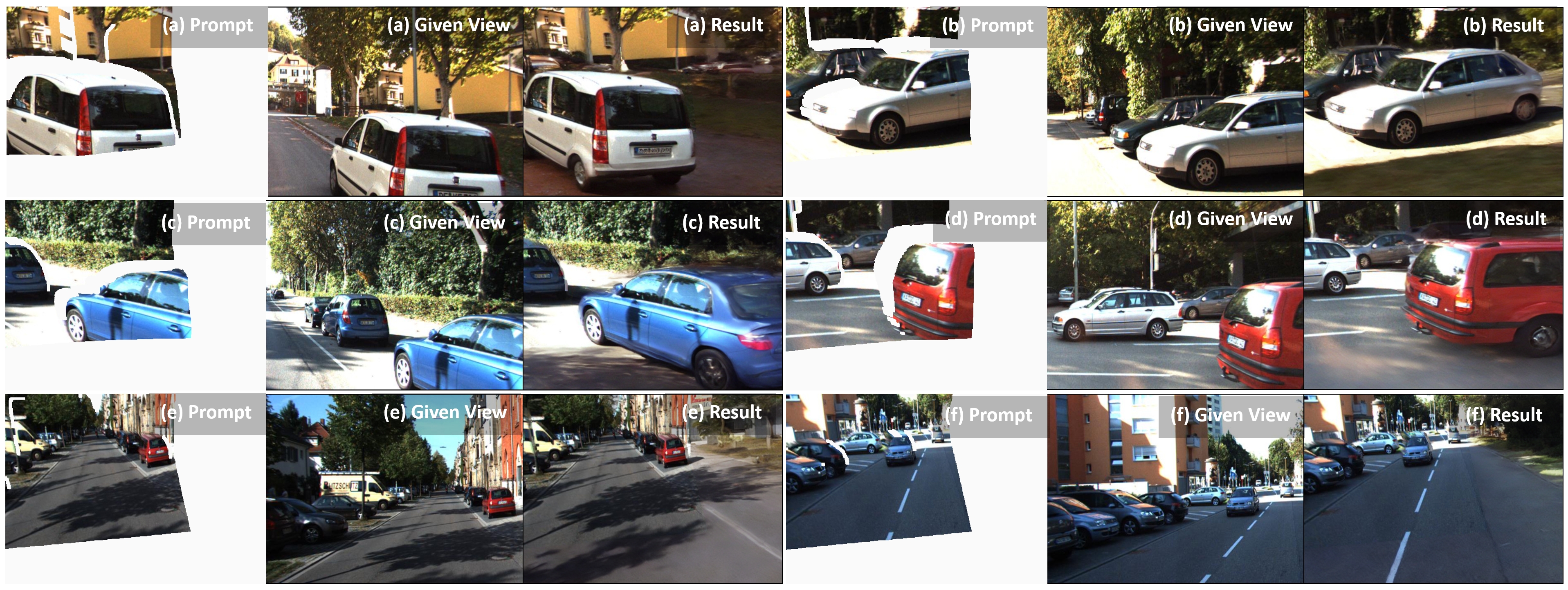}
    \caption{Novel view synthesis on KITTI dataset of 6 scenes from (\textbf{a}) to (\textbf{f}). For each scene, we give the "Prompt", which is warped from the "Given View", with the white regions referring to holes/missing regions. "Result" is our in-painting result from "Prompt". Note that for the synthesized novel view, there is no ground-truth available.} 
    \label{fig:exp-NVS}
    \vspace{-0.6cm}
\end{figure}

\subsection{Further analysis}
\label{Sec:further_analysis}

\begin{figure}[t]
   \centering
  \begin{minipage}[b]{0.32\textwidth}
    \centering
    \includegraphics[width=\textwidth]{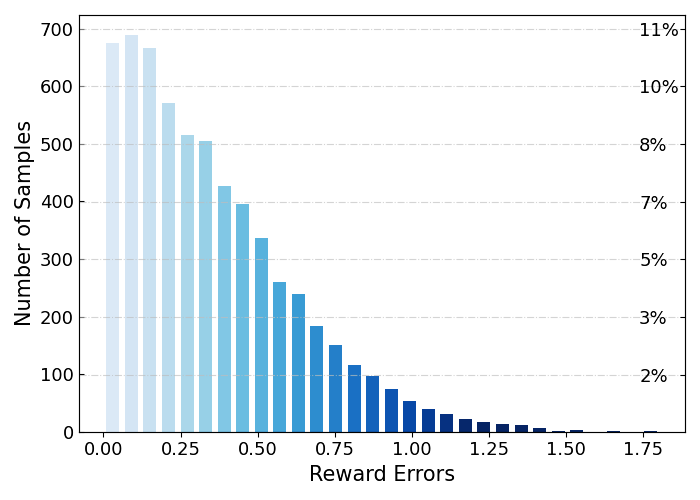}
    \vspace{-0.55cm}
    \caption{\label{Fig:error_barmap}Reward error distributions of the proposed reward model. The distribution of reward error percentages is depicted on the y-axis to the right.}   
    \vspace{-0.2cm}
  \end{minipage}
  \hfill
  \begin{minipage}[b]{0.35\textwidth}
    \centering
    \includegraphics[width=\textwidth]{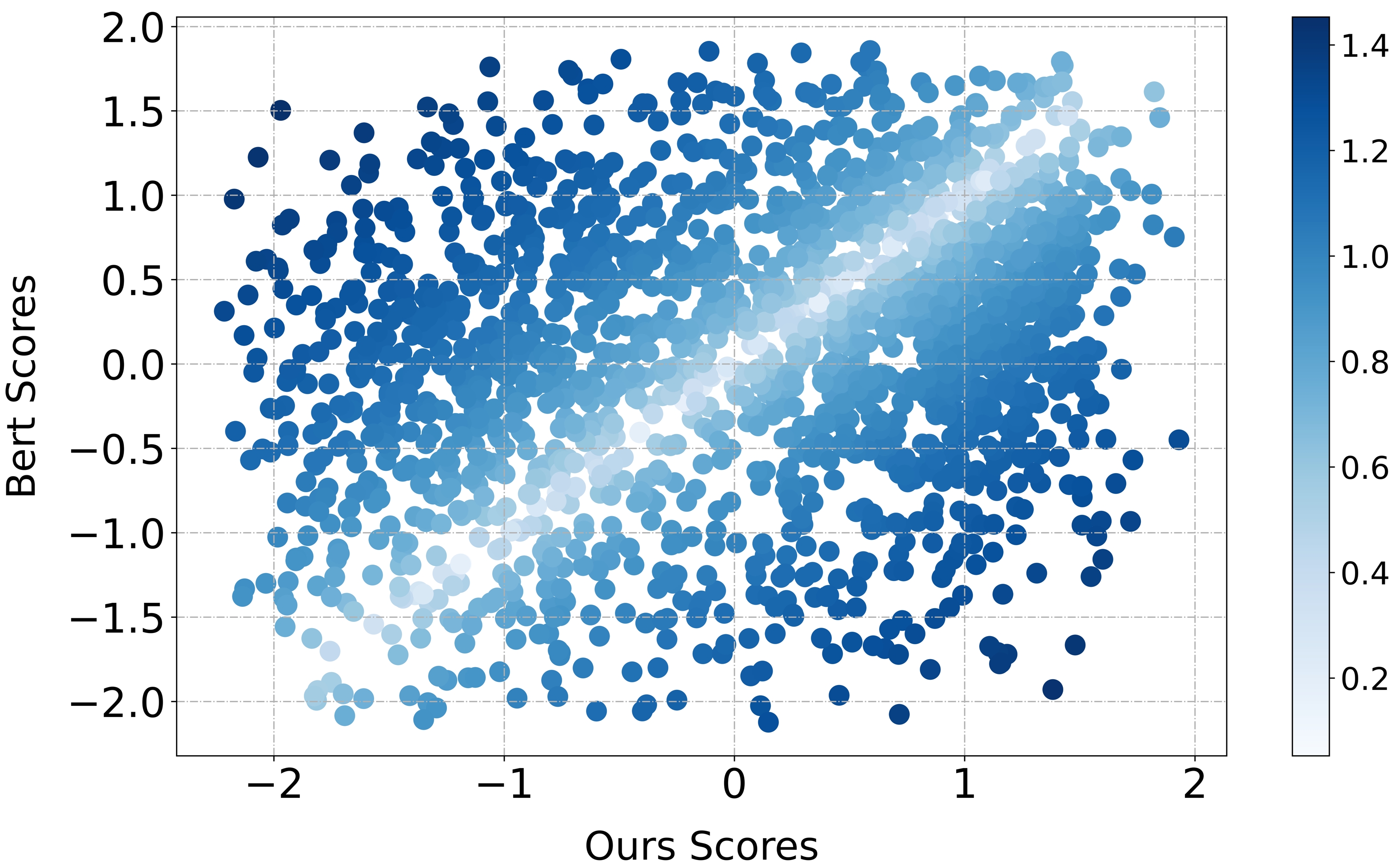}
    \vspace{-0.5cm}
    \caption{\label{Fig:error_scattermap}Plot of GT reward by our dataset (x-axis) and the Bert Score (y-axis) on the validation set, where each point indicates a sample. The darker indacated the larger error.}   
    \vspace{-0.2cm}
  \end{minipage}
  \hfill
  \begin{minipage}[b]{0.31\textwidth}
    \centering
    \includegraphics[width=\textwidth]{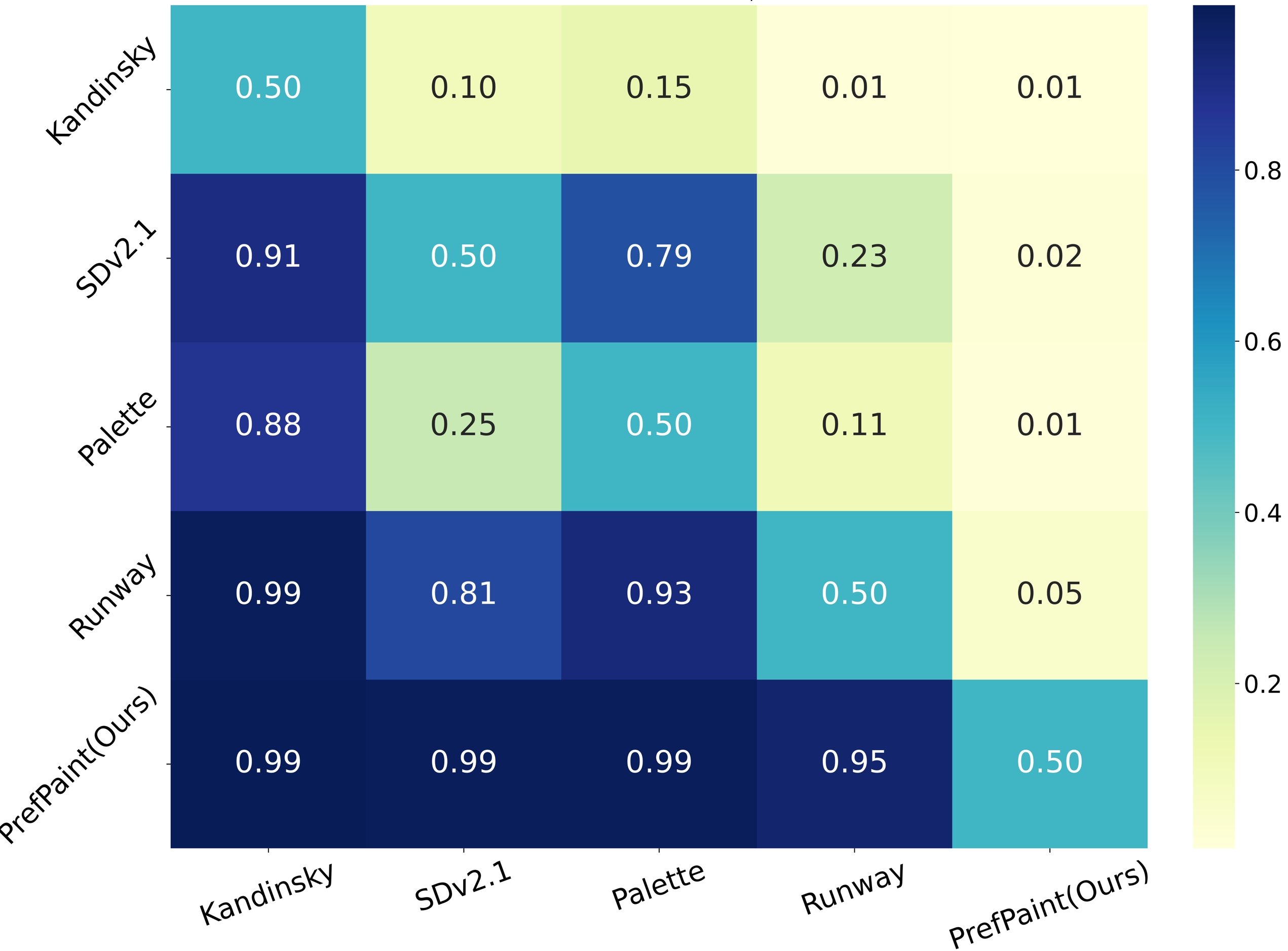}
    \vspace{-0.58cm}
    \caption{\label{Fig:WinRate_usrstudy}WinRate heat-map of user study. The winrate at specific locations shows the ratio of superior top methods to those at the bottom.}   
    \vspace{-0.2cm}
  \end{minipage}
    \vspace{-0.5cm}
\end{figure}

\noindent \textbf{Reward errors distribution.} We make statistics of reward estimation errors, and the results are shown in Fig.~\ref{Fig:error_barmap}. Although the proportion of very large error samples is not large, the incremental performance of our method lies in a more suitable choice of amplification function, as evidenced by Table~\ref{Tab_exp_functions}.

\noindent \textbf{Necessity of Our dataset.} Although existing metrics such as BERT Score~\cite{black2023training} provide a general measure of quality, we emphasize that our dataset is specifically tailored for the task of image inpainting, where human-labeled scores are both essential and more precise. To substantiate this claim, we conduct a comparison between the BERT score and our dataset’s score, as illustrated in Fig.~\ref{Fig:error_scattermap}. The results reveal a significant divergence between human judgments and BERT's preferences, underscoring the necessity and superior accuracy of the proposed dataset for this specific task.

\vspace{-0.2cm}
\subsection{Applications of Image FOV Enlargement and Novel View Synthesis}
\vspace{-0.3cm}
\label{Sec:Applications}
We also applied our approach to two additional tasks: (1) image field of view (FOV) enlargement, where we iteratively extended the boundaries of a typical image strip in the horizontal direction to create a  wider FOV image; and (2) novel view synthesis, where we warped a given image using the predicted depth image through the method in~\cite{yang2024depth} to generate a novel viewpoint and subsequently applied diffusion models to fill the holes/missing regions of the Warped view.  As depicted in Fig.~\ref{fig:exp-FOV} (\textcolor{red}{a})-a oil painting and (\textcolor{red}{b})-a realistic photography, our method yields natural and visually pleasing results that can be seamlessly integrated with the prompt regions. 
As illustrated in Figs.~\ref{fig:exp-NVS}, our method can generate more reasonable novel views and a visually appealing reconstruction.
We refer readers to the \textit{Supplementary Material} for more details about more application visual demonstrations.

\vspace{-0.3cm}
\section{Conclusion and Discussion}
\vspace{-0.4cm}
We have presented PrefPaint, an innovative scheme that leverages the principles of the linear bandit theorem to align diffusion models for image inpainting with human preferences iteratively. 
By integrating human feedback directly into the training loop, we have established a dynamic framework that not only respects the subjective nature of visual aesthetics but also continuously adapts to the evolving preferences of users. We conducted extensive experiments to demonstrate the necessity for alignment in the task of image inpainting and the significant superiority of PrefPaint both quantitatively and qualitatively. We believe that our method, along with the newly constructed dataset, has the potential to bring significant benefits to the development of visually driven AI applications.

PrefPaint aligns with a distribution that corresponds to the preferences of a specific group. However, it is important to recognize that individual preferences for image styles vary. Therefore, after achieving alignment with the general preferences of a group, it is advisable to develop personalized rewards and a corresponding reinforced alignment model to ensure complete alignment with the preferences of each user. Recently, the implementation of a reward-free alignment process, as discussed in~\cite{rafailov2024direct}, has gained popularity. Consequently, exploring reward model-free training methods for alignment in diffusion-based models represents a promising avenue for future research. Furthermore, there is an opportunity to explore the potential applications of our in-painting algorithm for additional 3D reconstruction tasks.

\clearpage
\section*{Acknowledgement}
This work was supported in part by the National Natural Science Foundation of China Excellent Young Scientists Fund 62422118, and in part by the Hong Kong Research Grants Council under Grant 11219324 and 11219422, and in part by the Hong Kong University Grants Council under Grant UGC/FDS11/E02/22.

\bibliography{main}

\begin{thebibliography}{10}
\providecommand{\url}[1]{#1}
\csname url@samestyle\endcsname
\providecommand{\newblock}{\relax}
\providecommand{\bibinfo}[2]{#2}
\providecommand{\BIBentrySTDinterwordspacing}{\spaceskip=0pt\relax}
\providecommand{\BIBentryALTinterwordstretchfactor}{4}
\providecommand{\BIBentryALTinterwordspacing}{\spaceskip=\fontdimen2\font plus
\BIBentryALTinterwordstretchfactor\fontdimen3\font minus \fontdimen4\font\relax}
\providecommand{\BIBforeignlanguage}[2]{{%
\expandafter\ifx\csname l@#1\endcsname\relax
\typeout{** WARNING: IEEEtran.bst: No hyphenation pattern has been}%
\typeout{** loaded for the language `#1'. Using the pattern for}%
\typeout{** the default language instead.}%
\else
\language=\csname l@#1\endcsname
\fi
#2}}
\providecommand{\BIBdecl}{\relax}
\BIBdecl

\bibitem{hou2024global}
J.~Hou, Z.~Zhu, J.~Hou, H.~Liu, H.~Zeng, and H.~Yuan, ``Global structure-aware diffusion process for low-light image enhancement,'' \emph{Advances in Neural Information Processing Systems}, vol.~36, 2024.

\bibitem{you2024nvs}
M.~You, Z.~Zhu, H.~Liu, and J.~Hou, ``Nvs-solver: Video diffusion model as zero-shot novel view synthesizer,'' \emph{arXiv preprint arXiv:2405.15364}, 2024.

\bibitem{black2023training}
K.~Black, M.~Janner, Y.~Du, I.~Kostrikov, and S.~Levine, ``Training diffusion models with reinforcement learning,'' \emph{arXiv preprint arXiv:2305.13301}, 2023.

\bibitem{ho2020denoising}
J.~Ho, A.~Jain, and P.~Abbeel, ``Denoising diffusion probabilistic models,'' \emph{Advances in neural information processing systems}, vol.~33, pp. 6840--6851, 2020.

\bibitem{song2020denoising}
J.~Song, C.~Meng, and S.~Ermon, ``Denoising diffusion implicit models,'' \emph{arXiv preprint arXiv:2010.02502}, 2020.

\bibitem{zhou2017scene}
B.~Zhou, H.~Zhao, X.~Puig, S.~Fidler, A.~Barriuso, and A.~Torralba, ``Scene parsing through ade20k dataset,'' in \emph{Proceedings of the IEEE conference on computer vision and pattern recognition}, 2017, pp. 633--641.

\bibitem{fan2024reinforcement}
Y.~Fan, O.~Watkins, Y.~Du, H.~Liu, M.~Ryu, C.~Boutilier, P.~Abbeel, M.~Ghavamzadeh, K.~Lee, and K.~Lee, ``Reinforcement learning for fine-tuning text-to-image diffusion models,'' \emph{Advances in Neural Information Processing Systems}, vol.~36, 2024.

\bibitem{zhou2019semantic}
B.~Zhou, H.~Zhao, X.~Puig, T.~Xiao, S.~Fidler, A.~Barriuso, and A.~Torralba, ``Semantic understanding of scenes through the ade20k dataset,'' \emph{International Journal of Computer Vision}, vol. 127, pp. 302--321, 2019.

\bibitem{Timofte_2018_CVPR_Workshops}
R.~Timofte, S.~Gu, J.~Wu, L.~Van~Gool, L.~Zhang, M.-H. Yang, M.~Haris \emph{et~al.}, ``Ntire 2018 challenge on single image super-resolution: Methods and results,'' in \emph{The IEEE Conference on Computer Vision and Pattern Recognition (CVPR) Workshops}, June 2018.

\bibitem{Agustsson_2017_CVPR_Workshops}
E.~Agustsson and R.~Timofte, ``Ntire 2017 challenge on single image super-resolution: Dataset and study,'' in \emph{The IEEE Conference on Computer Vision and Pattern Recognition (CVPR) Workshops}, July 2017.

\bibitem{Geiger2012CVPR}
A.~Geiger, P.~Lenz, and R.~Urtasun, ``Are we ready for autonomous driving? the kitti vision benchmark suite,'' in \emph{Conference on Computer Vision and Pattern Recognition (CVPR)}, 2012.

\bibitem{deng2009imagenet}
J.~Deng, W.~Dong, R.~Socher, L.-J. Li, K.~Li, and L.~Fei-Fei, ``Imagenet: A large-scale hierarchical image database,'' in \emph{2009 IEEE conference on computer vision and pattern recognition}.\hskip 1em plus 0.5em minus 0.4em\relax Ieee, 2009, pp. 248--255.

\bibitem{schulman2015trust}
J.~Schulman, S.~Levine, P.~Abbeel, M.~Jordan, and P.~Moritz, ``Trust region policy optimization,'' in \emph{International conference on machine learning}.\hskip 1em plus 0.5em minus 0.4em\relax PMLR, 2015, pp. 1889--1897.

\bibitem{kakade2001natural}
S.~M. Kakade, ``A natural policy gradient,'' \emph{Advances in neural information processing systems}, vol.~14, 2001.

\bibitem{yang2024depth}
L.~Yang, B.~Kang, Z.~Huang, X.~Xu, J.~Feng, and H.~Zhao, ``Depth anything: Unleashing the power of large-scale unlabeled data,'' \emph{arXiv preprint arXiv:2401.10891}, 2024.

\bibitem{li2020recurrent}
J.~Li, N.~Wang, L.~Zhang, B.~Du, and D.~Tao, ``Recurrent feature reasoning for image inpainting,'' in \emph{Proceedings of the IEEE/CVF conference on computer vision and pattern recognition}, 2020, pp. 7760--7768.

\bibitem{yi2020contextual}
Z.~Yi, Q.~Tang, S.~Azizi, D.~Jang, and Z.~Xu, ``Contextual residual aggregation for ultra high-resolution image inpainting,'' in \emph{Proceedings of the IEEE/CVF conference on computer vision and pattern recognition}, 2020, pp. 7508--7517.

\bibitem{anciukevivcius2023renderdiffusion}
T.~Anciukevi{\v{c}}ius, Z.~Xu, M.~Fisher, P.~Henderson, H.~Bilen, N.~J. Mitra, and P.~Guerrero, ``Renderdiffusion: Image diffusion for 3d reconstruction, inpainting and generation,'' in \emph{Proceedings of the IEEE/CVF Conference on Computer Vision and Pattern Recognition}, 2023, pp. 12\,608--12\,618.

\bibitem{svitov2023dinar}
D.~Svitov, D.~Gudkov, R.~Bashirov, and V.~Lempitsky, ``Dinar: Diffusion inpainting of neural textures for one-shot human avatars,'' in \emph{Proceedings of the IEEE/CVF International Conference on Computer Vision}, 2023, pp. 7062--7072.

\bibitem{kaelbling1996reinforcement}
L.~P. Kaelbling, M.~L. Littman, and A.~W. Moore, ``Reinforcement learning: A survey,'' \emph{Journal of artificial intelligence research}, vol.~4, pp. 237--285, 1996.

\bibitem{arulkumaran2017deep}
K.~Arulkumaran, M.~P. Deisenroth, M.~Brundage, and A.~A. Bharath, ``Deep reinforcement learning: A brief survey,'' \emph{IEEE Signal Processing Magazine}, vol.~34, no.~6, pp. 26--38, 2017.

\bibitem{franccois2018introduction}
V.~Fran{\c{c}}ois-Lavet, P.~Henderson, R.~Islam, M.~G. Bellemare, J.~Pineau \emph{et~al.}, ``An introduction to deep reinforcement learning,'' \emph{Foundations and Trends{\textregistered} in Machine Learning}, vol.~11, no. 3-4, pp. 219--354, 2018.

\bibitem{van2012reinforcement}
M.~Van~Otterlo and M.~Wiering, ``Reinforcement learning and markov decision processes,'' in \emph{Reinforcement learning: State-of-the-art}.\hskip 1em plus 0.5em minus 0.4em\relax Springer, 2012, pp. 3--42.

\bibitem{gattami2021reinforcement}
A.~Gattami, Q.~Bai, and V.~Aggarwal, ``Reinforcement learning for constrained markov decision processes,'' in \emph{International Conference on Artificial Intelligence and Statistics}.\hskip 1em plus 0.5em minus 0.4em\relax PMLR, 2021, pp. 2656--2664.

\bibitem{feinberg2012handbook}
E.~A. Feinberg and A.~Shwartz, \emph{Handbook of Markov decision processes: methods and applications}.\hskip 1em plus 0.5em minus 0.4em\relax Springer Science \& Business Media, 2012, vol.~40.

\bibitem{wang2020deep}
H.-n. Wang, N.~Liu, Y.-y. Zhang, D.-w. Feng, F.~Huang, D.-s. Li, and Y.-m. Zhang, ``Deep reinforcement learning: a survey,'' \emph{Frontiers of Information Technology \& Electronic Engineering}, vol.~21, no.~12, pp. 1726--1744, 2020.

\bibitem{zhang2024large}
Y.~Zhang, E.~Tzeng, Y.~Du, and D.~Kislyuk, ``Large-scale reinforcement learning for diffusion models,'' \emph{arXiv preprint arXiv:2401.12244}, 2024.

\bibitem{xu2024imagereward}
J.~Xu, X.~Liu, Y.~Wu, Y.~Tong, Q.~Li, M.~Ding, J.~Tang, and Y.~Dong, ``Imagereward: Learning and evaluating human preferences for text-to-image generation,'' \emph{Advances in Neural Information Processing Systems}, vol.~36, 2024.

\bibitem{ong2015distributed}
H.~Y. Ong, K.~Chavez, and A.~Hong, ``Distributed deep q-learning,'' \emph{arXiv preprint arXiv:1508.04186}, 2015.

\bibitem{mnih2013playing}
V.~Mnih, K.~Kavukcuoglu, D.~Silver, A.~Graves, I.~Antonoglou, D.~Wierstra, and M.~Riedmiller, ``Playing atari with deep reinforcement learning,'' \emph{arXiv preprint arXiv:1312.5602}, 2013.

\bibitem{mnih2015human}
V.~Mnih, K.~Kavukcuoglu, D.~Silver, A.~A. Rusu, J.~Veness, M.~G. Bellemare, A.~Graves, M.~Riedmiller, A.~K. Fidjeland, G.~Ostrovski \emph{et~al.}, ``Human-level control through deep reinforcement learning,'' \emph{nature}, vol. 518, no. 7540, pp. 529--533, 2015.

\bibitem{schulman2017proximal}
J.~Schulman, F.~Wolski, P.~Dhariwal, A.~Radford, and O.~Klimov, ``Proximal policy optimization algorithms,'' \emph{arXiv preprint arXiv:1707.06347}, 2017.

\bibitem{khadka2018evolution}
S.~Khadka and K.~Tumer, ``Evolution-guided policy gradient in reinforcement learning,'' \emph{Advances in Neural Information Processing Systems}, vol.~31, 2018.

\bibitem{lu2021decentralized}
S.~Lu, K.~Zhang, T.~Chen, T.~Ba{\c{s}}ar, and L.~Horesh, ``Decentralized policy gradient descent ascent for safe multi-agent reinforcement learning,'' in \emph{Proceedings of the AAAI Conference on Artificial Intelligence}, vol.~35, no.~10, 2021, pp. 8767--8775.

\bibitem{bahdanau2016actor}
D.~Bahdanau, P.~Brakel, K.~Xu, A.~Goyal, R.~Lowe, J.~Pineau, A.~Courville, and Y.~Bengio, ``An actor-critic algorithm for sequence prediction,'' \emph{arXiv preprint arXiv:1607.07086}, 2016.

\bibitem{fujimoto2018addressing}
S.~Fujimoto, H.~Hoof, and D.~Meger, ``Addressing function approximation error in actor-critic methods,'' in \emph{International conference on machine learning}.\hskip 1em plus 0.5em minus 0.4em\relax PMLR, 2018, pp. 1587--1596.

\bibitem{haarnoja2018soft}
T.~Haarnoja, A.~Zhou, K.~Hartikainen, G.~Tucker, S.~Ha, J.~Tan, V.~Kumar, H.~Zhu, A.~Gupta, P.~Abbeel \emph{et~al.}, ``Soft actor-critic algorithms and applications,'' \emph{arXiv preprint arXiv:1812.05905}, 2018.

\bibitem{shinn2024reflexion}
N.~Shinn, F.~Cassano, A.~Gopinath, K.~Narasimhan, and S.~Yao, ``Reflexion: Language agents with verbal reinforcement learning,'' \emph{Advances in Neural Information Processing Systems}, vol.~36, 2024.

\bibitem{rafailov2024direct}
R.~Rafailov, A.~Sharma, E.~Mitchell, C.~D. Manning, S.~Ermon, and C.~Finn, ``Direct preference optimization: Your language model is secretly a reward model,'' \emph{Advances in Neural Information Processing Systems}, vol.~36, 2024.

\bibitem{kasneci2023chatgpt}
E.~Kasneci, K.~Se{\ss}ler, S.~K{\"u}chemann, M.~Bannert, D.~Dementieva, F.~Fischer, U.~Gasser, G.~Groh, S.~G{\"u}nnemann, E.~H{\"u}llermeier \emph{et~al.}, ``Chatgpt for good? on opportunities and challenges of large language models for education,'' \emph{Learning and individual differences}, vol. 103, p. 102274, 2023.

\bibitem{liu2023gpt}
X.~Liu, Y.~Zheng, Z.~Du, M.~Ding, Y.~Qian, Z.~Yang, and J.~Tang, ``Gpt understands, too,'' \emph{AI Open}, 2023.

\bibitem{ruthotto2021introduction}
L.~Ruthotto and E.~Haber, ``An introduction to deep generative modeling,'' \emph{GAMM-Mitteilungen}, vol.~44, no.~2, p. e202100008, 2021.

\bibitem{sohl2015deep}
J.~Sohl-Dickstein, E.~Weiss, N.~Maheswaranathan, and S.~Ganguli, ``Deep unsupervised learning using nonequilibrium thermodynamics,'' in \emph{International conference on machine learning}.\hskip 1em plus 0.5em minus 0.4em\relax PMLR, 2015, pp. 2256--2265.

\bibitem{lugmayr2022repaint}
A.~Lugmayr, M.~Danelljan, A.~Romero, F.~Yu, R.~Timofte, and L.~Van~Gool, ``Repaint: Inpainting using denoising diffusion probabilistic models,'' in \emph{Proceedings of the IEEE/CVF Conference on Computer Vision and Pattern Recognition}, 2022, pp. 11\,461--11\,471.

\bibitem{corneanu2024latentpaint}
C.~Corneanu, R.~Gadde, and A.~M. Martinez, ``Latentpaint: Image inpainting in latent space with diffusion models,'' in \emph{Proceedings of the IEEE/CVF Winter Conference on Applications of Computer Vision}, 2024, pp. 4334--4343.

\bibitem{saharia2022palette}
C.~Saharia, W.~Chan, H.~Chang, C.~Lee, J.~Ho, T.~Salimans, D.~Fleet, and M.~Norouzi, ``Palette: Image-to-image diffusion models,'' in \emph{ACM SIGGRAPH 2022 Conference Proceedings}, 2022, pp. 1--10.

\bibitem{song2023consistency}
Y.~Song, P.~Dhariwal, M.~Chen, and I.~Sutskever, ``Consistency models,'' 2023.

\bibitem{song2020score}
Y.~Song, J.~Sohl-Dickstein, D.~P. Kingma, A.~Kumar, S.~Ermon, and B.~Poole, ``Score-based generative modeling through stochastic differential equations,'' \emph{arXiv preprint arXiv:2011.13456}, 2020.

\bibitem{stiennon2020learning}
N.~Stiennon, L.~Ouyang, J.~Wu, D.~Ziegler, R.~Lowe, C.~Voss, A.~Radford, D.~Amodei, and P.~F. Christiano, ``Learning to summarize with human feedback,'' \emph{Advances in Neural Information Processing Systems}, vol.~33, pp. 3008--3021, 2020.

\bibitem{casper2023open}
S.~Casper, X.~Davies, C.~Shi, T.~K. Gilbert, J.~Scheurer, J.~Rando, R.~Freedman, T.~Korbak, D.~Lindner, P.~Freire \emph{et~al.}, ``Open problems and fundamental limitations of reinforcement learning from human feedback,'' \emph{arXiv preprint arXiv:2307.15217}, 2023.

\bibitem{munos2023nash}
R.~Munos, M.~Valko, D.~Calandriello, M.~G. Azar, M.~Rowland, Z.~D. Guo, Y.~Tang, M.~Geist, T.~Mesnard, A.~Michi \emph{et~al.}, ``Nash learning from human feedback,'' \emph{arXiv preprint arXiv:2312.00886}, 2023.

\bibitem{stypulkowski2024diffused}
M.~Stypu{\l}kowski, K.~Vougioukas, S.~He, M.~Zieba, S.~Petridis, and M.~Pantic, ``Diffused heads: Diffusion models beat gans on talking-face generation,'' in \emph{Proceedings of the IEEE/CVF Winter Conference on Applications of Computer Vision}, 2024, pp. 5091--5100.

\bibitem{muller2022diffusion}
G.~M{\"u}ller-Franzes, J.~M. Niehues, F.~Khader, S.~T. Arasteh, C.~Haarburger, C.~Kuhl, T.~Wang, T.~Han, S.~Nebelung, J.~N. Kather \emph{et~al.}, ``Diffusion probabilistic models beat gans on medical images,'' \emph{arXiv preprint arXiv:2212.07501}, 2022.

\bibitem{ho2022imagen}
J.~Ho, W.~Chan, C.~Saharia, J.~Whang, R.~Gao, A.~Gritsenko, D.~P. Kingma, B.~Poole, M.~Norouzi, D.~J. Fleet \emph{et~al.}, ``Imagen video: High definition video generation with diffusion models,'' \emph{arXiv preprint arXiv:2210.02303}, 2022.

\bibitem{karaca2016interpolation}
E.~Karaca and M.~A. Tunga, ``Interpolation-based image inpainting in color images using high dimensional model representation,'' in \emph{2016 24th European Signal Processing Conference (EUSIPCO)}.\hskip 1em plus 0.5em minus 0.4em\relax IEEE, 2016, pp. 2425--2429.

\bibitem{arias2012nonlocal}
P.~Arias, V.~Caselles, G.~Facciolo, V.~Lazcano, and R.~Sadek, ``Nonlocal variational models for inpainting and interpolation,'' \emph{Mathematical Models and Methods in Applied Sciences}, vol.~22, no. supp02, p. 1230003, 2012.

\bibitem{alsalamah2016medical}
M.~Alsalamah and S.~Amin, ``Medical image inpainting with rbf interpolation technique,'' \emph{International Journal of Advanced Computer Science and Applications}, vol.~7, no.~8, 2016.

\bibitem{guo2021image}
X.~Guo, H.~Yang, and D.~Huang, ``Image inpainting via conditional texture and structure dual generation,'' in \emph{Proceedings of the IEEE/CVF International Conference on Computer Vision}, 2021, pp. 14\,134--14\,143.

\bibitem{jain2023keys}
J.~Jain, Y.~Zhou, N.~Yu, and H.~Shi, ``Keys to better image inpainting: Structure and texture go hand in hand,'' in \emph{Proceedings of the IEEE/CVF Winter Conference on Applications of Computer Vision}, 2023, pp. 208--217.

\bibitem{lahiri2020prior}
A.~Lahiri, A.~K. Jain, S.~Agrawal, P.~Mitra, and P.~K. Biswas, ``Prior guided gan based semantic inpainting,'' in \emph{Proceedings of the IEEE/CVF conference on computer vision and pattern recognition}, 2020, pp. 13\,696--13\,705.

\bibitem{zhang2022gan}
X.~Zhang, X.~Wang, C.~Shi, Z.~Yan, X.~Li, B.~Kong, S.~Lyu, B.~Zhu, J.~Lv, Y.~Yin \emph{et~al.}, ``De-gan: Domain embedded gan for high quality face image inpainting,'' \emph{Pattern Recognition}, vol. 124, p. 108415, 2022.

\bibitem{yang2023uni}
S.~Yang, X.~Chen, and J.~Liao, ``Uni-paint: A unified framework for multimodal image inpainting with pretrained diffusion model,'' in \emph{Proceedings of the 31st ACM International Conference on Multimedia}, 2023, pp. 3190--3199.

\bibitem{wu2023human}
X.~Wu, K.~Sun, F.~Zhu, R.~Zhao, and H.~Li, ``Human preference score: Better aligning text-to-image models with human preference,'' in \emph{Proceedings of the IEEE/CVF International Conference on Computer Vision}, 2023, pp. 2096--2105.

\bibitem{satgunam2013factors}
P.~N. Satgunam, R.~L. Woods, P.~M. Bronstad, and E.~Peli, ``Factors affecting enhanced video quality preferences,'' \emph{IEEE transactions on image processing}, vol.~22, no.~12, pp. 5146--5157, 2013.

\bibitem{abbasi2011improved}
Y.~Abbasi-Yadkori, D.~P{\'a}l, and C.~Szepesv{\'a}ri, ``Improved algorithms for linear stochastic bandits,'' \emph{Advances in neural information processing systems}, vol.~24, 2011.

\bibitem{kandinsky}
A.~Razzhigaev, A.~Shakhmatov, A.~Maltseva, V.~Arkhipkin, I.~Pavlov, I.~Ryabov, A.~Kuts, A.~Panchenko, A.~Kuznetsov, and D.~Dimitrov, ``Kandinsky: an improved text-to-image synthesis with image prior and latent diffusion,'' \emph{arXiv preprint arXiv:2310.03502}, 2023.

\bibitem{sd1.5}
\BIBentryALTinterwordspacing
Runway, ``Runway model on huggingface,'' 2023. [Online]. Available: \url{https://huggingface.co/runwayml/stable-diffusion-v1-5}
\BIBentrySTDinterwordspacing

\bibitem{sd2.1}
\BIBentryALTinterwordspacing
StabilityAI, ``Stabilityai model on huggingface,'' 2023. [Online]. Available: \url{https://huggingface.co/stabilityai/stable-diffusion-2-1}
\BIBentrySTDinterwordspacing

\bibitem{compvis}
\BIBentryALTinterwordspacing
Compvis, ``Compvis github website,'' 2023. [Online]. Available: \url{https://github.com/CompVis/latent-diffusion}
\BIBentrySTDinterwordspacing

\bibitem{sdxl}
\BIBentryALTinterwordspacing
StabilityAI, ``Stabilityai model on huggingface,'' 2023. [Online]. Available: \url{https://huggingface.co/stabilityai/stable-diffusion-xl-base-1.0}
\BIBentrySTDinterwordspacing

\bibitem{sdxlInpt}
\BIBentryALTinterwordspacing
Diffuser, ``Stabilityai model on huggingface,'' 2023. [Online]. Available: \url{https://huggingface.co/diffusers/stable-diffusion-xl-1.0-inpainting-0.1}
\BIBentrySTDinterwordspacing

\bibitem{runway}
\BIBentryALTinterwordspacing
Runway, ``Runway model on huggingface,'' 2023. [Online]. Available: \url{https://huggingface.co/runwayml/stable-diffusion-inpainting}
\BIBentrySTDinterwordspacing

\bibitem{MAT}
W.~Li, Z.~Lin, K.~Zhou, L.~Qi, Y.~Wang, and J.~Jia, ``Mat: Mask-aware transformer for large hole image inpainting,'' in \emph{Proceedings of the IEEE/CVF conference on computer vision and pattern recognition}, 2022, pp. 10\,758--10\,768.

\bibitem{clip}
A.~Radford, J.~W. Kim, C.~Hallacy, A.~Ramesh, G.~Goh, S.~Agarwal, G.~Sastry, A.~Askell, P.~Mishkin, J.~Clark \emph{et~al.}, ``Learning transferable visual models from natural language supervision,'' in \emph{International conference on machine learning}.\hskip 1em plus 0.5em minus 0.4em\relax PMLR, 2021, pp. 8748--8763.

\bibitem{blip}
J.~Li, D.~Li, C.~Xiong, and S.~Hoi, ``Blip: Bootstrapping language-image pre-training for unified vision-language understanding and generation,'' in \emph{International conference on machine learning}.\hskip 1em plus 0.5em minus 0.4em\relax PMLR, 2022, pp. 12\,888--12\,900.

\bibitem{schuhmann2022laion}
C.~Schuhmann, R.~Beaumont, R.~Vencu, C.~Gordon, R.~Wightman, M.~Cherti, T.~Coombes, A.~Katta, C.~Mullis, M.~Wortsman \emph{et~al.}, ``Laion-5b: An open large-scale dataset for training next generation image-text models,'' \emph{Advances in Neural Information Processing Systems}, vol.~35, pp. 25\,278--25\,294, 2022.

\bibitem{szegedy2016rethinking}
C.~Szegedy, V.~Vanhoucke, S.~Ioffe, J.~Shlens, and Z.~Wojna, ``Rethinking the inception architecture for computer vision,'' in \emph{Proceedings of the IEEE conference on computer vision and pattern recognition}, 2016, pp. 2818--2826.

\bibitem{salimans2017pixelcnn++}
T.~Salimans, A.~Karpathy, X.~Chen, and D.~P. Kingma, ``Pixelcnn++: Improving the pixelcnn with discretized logistic mixture likelihood and other modifications,'' \emph{arXiv preprint arXiv:1701.05517}, 2017.

\bibitem{xiang20233d}
J.~Xiang, J.~Yang, B.~Huang, and X.~Tong, ``3d-aware image generation using 2d diffusion models,'' in \emph{Proceedings of the IEEE/CVF International Conference on Computer Vision}, 2023, pp. 2383--2393.

\bibitem{powerpaint}
J.~Zhuang, Y.~Zeng, W.~Liu, C.~Yuan, and K.~Chen, ``A task is worth one word: Learning with task prompts for high-quality versatile image inpainting,'' \emph{arXiv preprint arXiv:2312.03594}, 2023.

\bibitem{brushnet}
X.~Ju, X.~Liu, X.~Wang, Y.~Bian, Y.~Shan, and Q.~Xu, ``Brushnet: A plug-and-play image inpainting model with decomposed dual-branch diffusion,'' \emph{arXiv preprint arXiv:2403.06976}, 2024.

\bibitem{hdpaint}
H.~Manukyan, A.~Sargsyan, B.~Atanyan, Z.~Wang, S.~Navasardyan, and H.~Shi, ``Hd-painter: high-resolution and prompt-faithful text-guided image inpainting with diffusion models,'' \emph{arXiv preprint arXiv:2312.14091}, 2023.

\bibitem{yang2024using}
K.~Yang, J.~Tao, J.~Lyu, C.~Ge, J.~Chen, W.~Shen, X.~Zhu, and X.~Li, ``Using human feedback to fine-tune diffusion models without any reward model,'' in \emph{Proceedings of the IEEE/CVF Conference on Computer Vision and Pattern Recognition}, 2024, pp. 8941--8951.

\end{thebibliography}

\clearpage
\renewcommand \thepart{}
\renewcommand \partname{}

\appendix
\part{Appendix} 
\vspace{-1cm}

\setstretch{1.5}
\textcolor{black}{\parttoc} 

\setstretch{1}

\clearpage

\section{Dataset Details}

In this section, we visualize our labeling platform and the corresponding labeling scheme. 

\subsection{Labeling Scheme }
The reward score contains the following 3 terms:
\let\oldthefigure\thefigure
\let\oldthetable\thetable
\renewcommand{\thefigure}{\thesection-\oldthefigure}
\renewcommand{\thetable}{\thesection-\oldthetable}
\setcounter{figure}{0}   
\setcounter{table}{0}   

\textbf{Structural Rationality}. It measures the correctness of the model understanding of image content. 
\textbf{Score (0-2):} Failure generation, e.g. strange and weird objects; random patterns completed; inappropriate strings; abrupt bar completions. Objects are generated but obviously do not belong to the scene, e.g. There are shelves unique to homes on the side of the highway; there is a marble floor blocking the front of the washing machine. \textbf{Score (3-5):}The understanding of the scene is roughly correct, e.g. the inpainted sky next to the sky pattern; the sea behind the beach. \textbf{Score (6-7):} The overall structure of the generated scene is reasonable and do not seem to be any major problems.
To illustrate the scoring scheme, we show some examples and corresponding reasons in Table~\ref{Tab:Structural}.
\begin{table}[h]
  \centering
  \caption{\label{Tab:Structural}Illustration of different scoring examples on structural rationality.}
  \begin{tabular}{  p{0.45\columnwidth} | p{0.45\columnwidth}  }
    \hline\hline
      Prompt with Reconstruction &  Prompt with Reconstruction   \\ \hline
    \begin{minipage}[b]{0.45\columnwidth}
		\centering
		\raisebox{0pt}{\includegraphics[width=\linewidth]{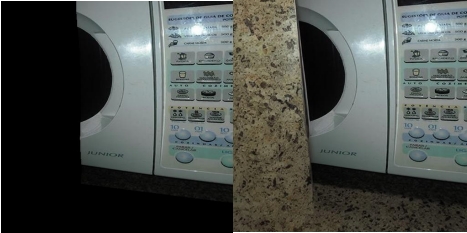}}
	\end{minipage}
    & 
    \begin{minipage}[b]{0.45\columnwidth}
		\centering
		\raisebox{0pt}{\includegraphics[width=\linewidth]{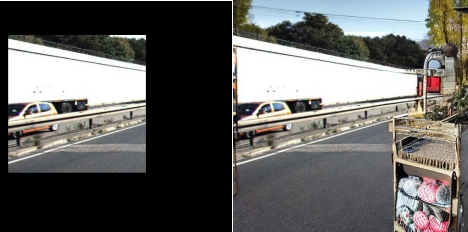}}
	\end{minipage}\\
 \textbf{Score:} 0. \textbf{Reason:} there should be no marble in front of the washing machine. Wrong semantics and wrong understanding of the scene.
    & \textbf{Score:} 1. \textbf{Reason:} the appearance of racks on the highway is unreasonable, semantically incorrect, and misunderstanding of the scene.\\\hline
    \begin{minipage}[b]{0.45\columnwidth}
		\centering
		\raisebox{0pt}{\includegraphics[width=\linewidth]{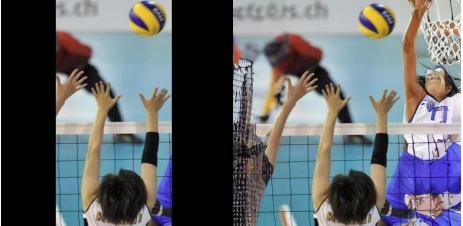}}
	\end{minipage} &
    \begin{minipage}[b]{0.45\columnwidth}
		\centering
		\raisebox{0pt}{\includegraphics[width=\linewidth]{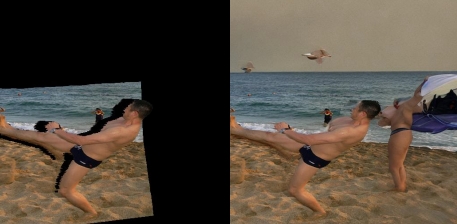}}
	\end{minipage}\\
 
     \textbf{Score:} 3. \textbf{Reason:} the model understands that this is a volleyball match and try to complete relevant people. &
    \textbf{Score:} 4. \textbf{Reason:} Although the quality of the details generated here is not rated high, the model understands that the scene is on the beach, and the overall structure is rated.\\\hline
    \begin{minipage}[b]{0.45\columnwidth}
		\centering
		\raisebox{0pt}{\includegraphics[width=\linewidth]{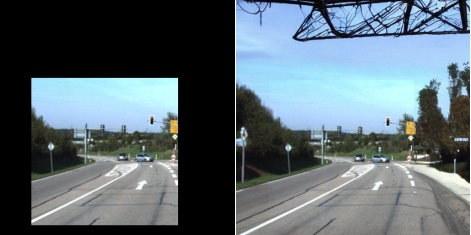}}
	\end{minipage}
    & \begin{minipage}[b]{0.45\columnwidth}
		\centering
		\raisebox{0pt}{\includegraphics[width=\linewidth]{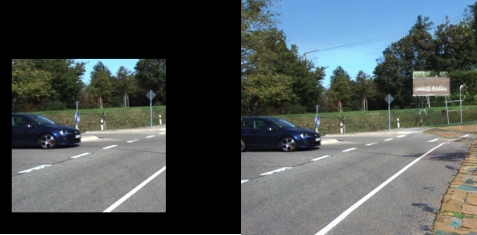}}
	\end{minipage}\\
    \textbf{Score:} 6. \textbf{Reason:} although there are some flaws, I understand the content of the highway scene and complete the relevant content. & \textbf{Score:} 7. \textbf{Reason:} Understand the scenario and correctly add relevant content.\\\hline\hline
  \end{tabular}
\end{table}

\textbf{Feeling of Local Texture.} It measures whether the texture of the object is correct and consistent with objective facts. \textbf{Score (0-2):} Failure generation (e.g. strange and weird objects; random patterns completed; inappropriate strings; abrupt bar completions). The generated objects do not conform to objective facts (e.g. a dog has two heads; a person without an upper body). The texture of the generated objects is very poor (e.g. two car bodies stuck together; poor quality sky). \textbf{Score (3-5):} The generated objects are generally reasonable (e.g. a complete leopard, but with very long legs; a complete corn, but the corn kernels are strange; there are flaws in the complete object;). The generated texture is not obtrusive but will have flaws if you zoom in. \textbf{Score (6-7):} Details are complete and reasonable. We also show some examples and corresponding reason in Table~\ref{Tab:Details}.

\begin{table}[h]
  \centering
  \caption{\label{Tab:Details}Illustration of different scoring examples on the feeling of local texture.}
  \begin{tabular}{  p{0.45\columnwidth} | p{0.45\columnwidth}  }
    \hline\hline
      Prompt with Reconstruction &  Prompt with Reconstruction   \\ \hline
    \begin{minipage}[b]{0.45\columnwidth}
		\centering
		\raisebox{0pt}{\includegraphics[width=\linewidth]{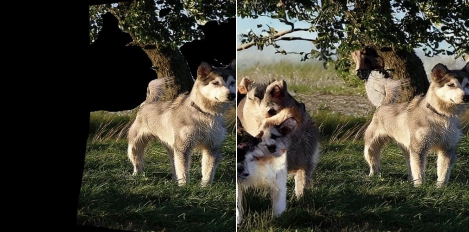}}
	\end{minipage}
    & 
    \begin{minipage}[b]{0.45\columnwidth}
		\centering
		\raisebox{0pt}{\includegraphics[width=\linewidth]{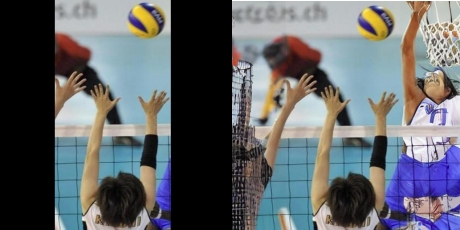}}
	\end{minipage}\\
 \textbf{Score:} 0. \textbf{Reason:} it can be seen that the model understands the scene correctly and wants to add a dog, but the quality is not high and does not conform to the objective facts.
    & \textbf{Score:} 1. \textbf{Reason:} it wants to generate people but the generation quality is not high.\\\hline
    \begin{minipage}[b]{0.45\columnwidth}
		\centering
		\raisebox{0pt}{\includegraphics[width=\linewidth]{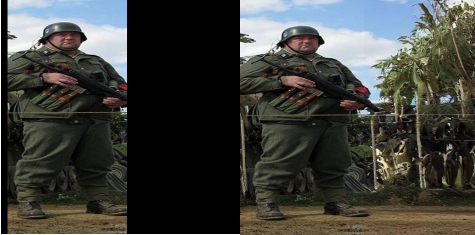}}
	\end{minipage} &
    \begin{minipage}[b]{0.45\columnwidth}
		\centering
		\raisebox{0pt}{\includegraphics[width=\linewidth]{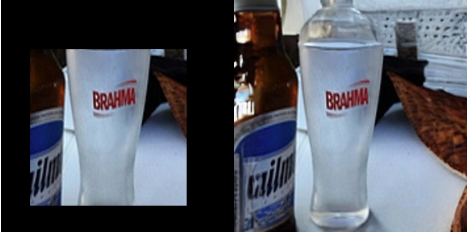}}
	\end{minipage}\\
 
     \textbf{Score:} 4. \textbf{Reason:} The completion is generally reasonable. &
    \textbf{Score:} 4. \textbf{Reason:} The completion is generally reasonable.\\\hline
    \begin{minipage}[b]{0.45\columnwidth}
		\centering
		\raisebox{0pt}{\includegraphics[width=\linewidth]{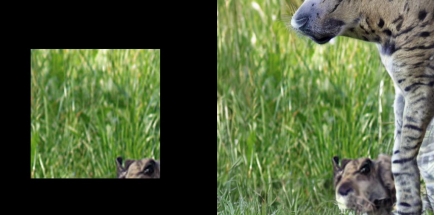}}
	\end{minipage}
    & \begin{minipage}[b]{0.45\columnwidth}
		\centering
		\raisebox{0pt}{\includegraphics[width=\linewidth]{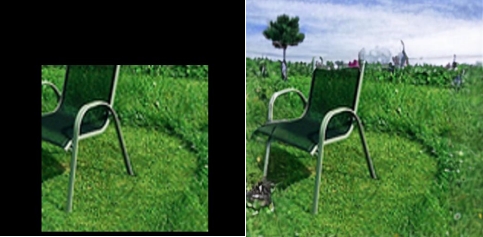}}
	\end{minipage}\\
    \textbf{Score:} 6. \textbf{Reason:} it wants to patch up the leopard but the legs is not consistent with common sense, but it can also be seen that it is a complete leopard. & \textbf{Score:} 7. \textbf{Reason:} The completion quality is good. good.\\\hline\hline
  \end{tabular}
\end{table}

\textbf{Overall Feeling.} The overall feeling given by the picture, whether it is reasonable and consistent with the objective facts. We directly show some examples and corresponding reasons in Table~\ref{Tab:Overall}.

\begin{table}[htbp]
  \centering
  \caption{\label{Tab:Overall}Illustration of different scoring examples on the overall feeling.}
  \begin{tabular}{  p{0.9\columnwidth}    }
    \hline\hline
      Prompt with Reconstruction (from left to right are Recon-1, 2, 3 and the prompt )   \\ \hline
    \begin{minipage}[b]{0.9\columnwidth}
		\centering
		\raisebox{0pt}{\includegraphics[width=0.9\columnwidth]{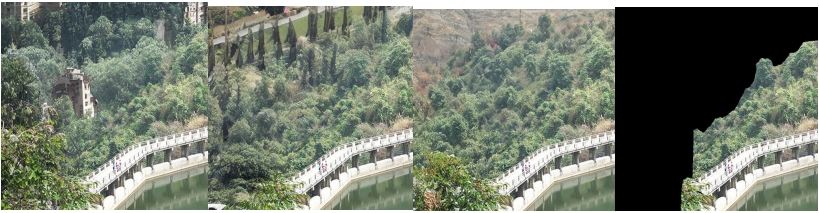}}
	\end{minipage}\\
 \textbf{Score:} 3, 2, 6. \textbf{Reason:} The quality of Recon-3 is significantly better than Recon-1, and there are no obvious flaws, so it has the highest score of 6.\\ \hline
    \begin{minipage}[b]{0.9\columnwidth}
		\centering
		\raisebox{0pt}{\includegraphics[width=0.9\columnwidth]{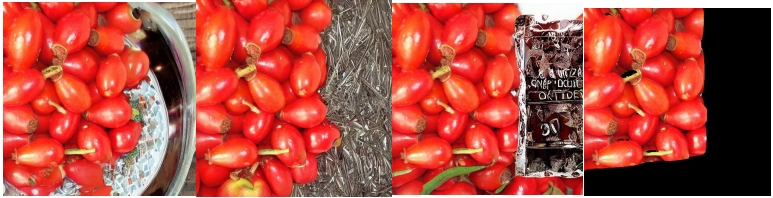}}
	\end{minipage}\\
 \textbf{Score:} 5, 3, 1. \textbf{Reason:} Recon-1 is the most reasonable, but the steel basin structure is a bit strange. Recon-2 may want to make up for the melon seeds. The details are quite good but the scenes are relatively rare. Recon-3 is very strange.\\ \hline
    \begin{minipage}[b]{0.9\columnwidth}
		\centering
		\raisebox{0pt}{\includegraphics[width=0.9\columnwidth]{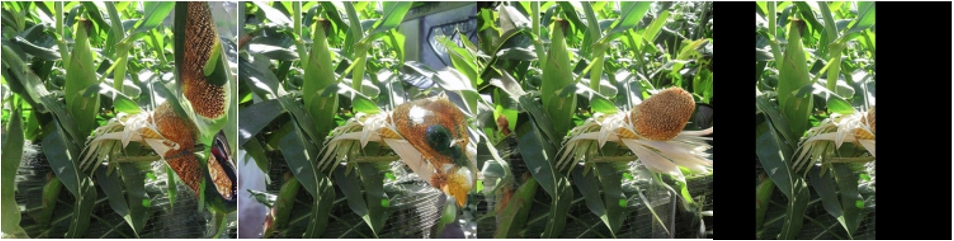}}
	\end{minipage}\\
 \textbf{Score:} 1, 1, 4. \textbf{Reason:} The overall structure of Recon-1 is wrong, and the quality of the corn heads in Recon-2 is relatively poor, and it looks like there is glue stuck on it. Recon-3 has a reasonable structure, but the proportion of the head is relatively large, and the grains of the corn head are strange, so the score of only 4 is not very high.\\ \hline
    \begin{minipage}[b]{0.9\columnwidth}
		\centering
		\raisebox{0pt}{\includegraphics[width=0.9\columnwidth]{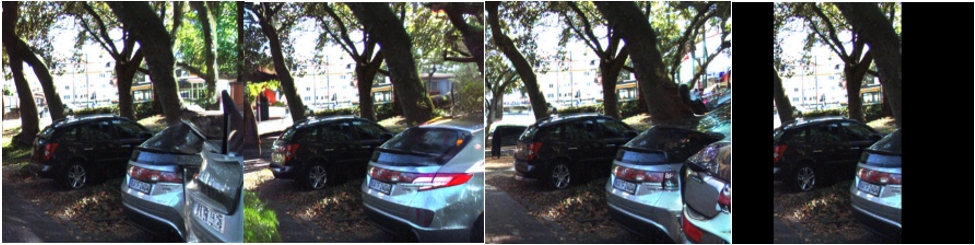}}
	\end{minipage}\\
 \textbf{Score:} 1, 7, 2. \textbf{Reason:} The shape of the car on the right side of Recon-1 is too weird and affects the look and feel. Recon-2 is of very good quality and very natural. The house on the right side of Recon-3 has a poorer perspective.\\ \hline
    \begin{minipage}[b]{0.9\columnwidth}
		\centering
		\raisebox{0pt}{\includegraphics[width=0.9\columnwidth]{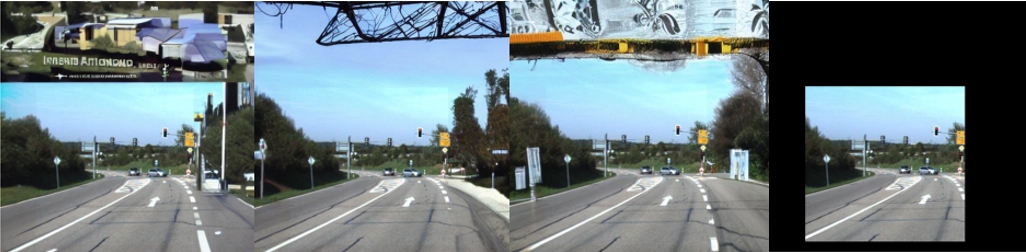}}
	\end{minipage}\\
 \textbf{Score:} 0, 4, 1. \textbf{Reason:} Recon-1 was randomly generated, Recon-3 tried to generate a traffic light, but failed. Similar branches in Recon-2 don’t look so inconsistent.\\
    \hline\hline
  \end{tabular}
\end{table}

\begin{figure}[h]
  \centering
  \resizebox{1.0\textwidth}{!}{\includegraphics[]{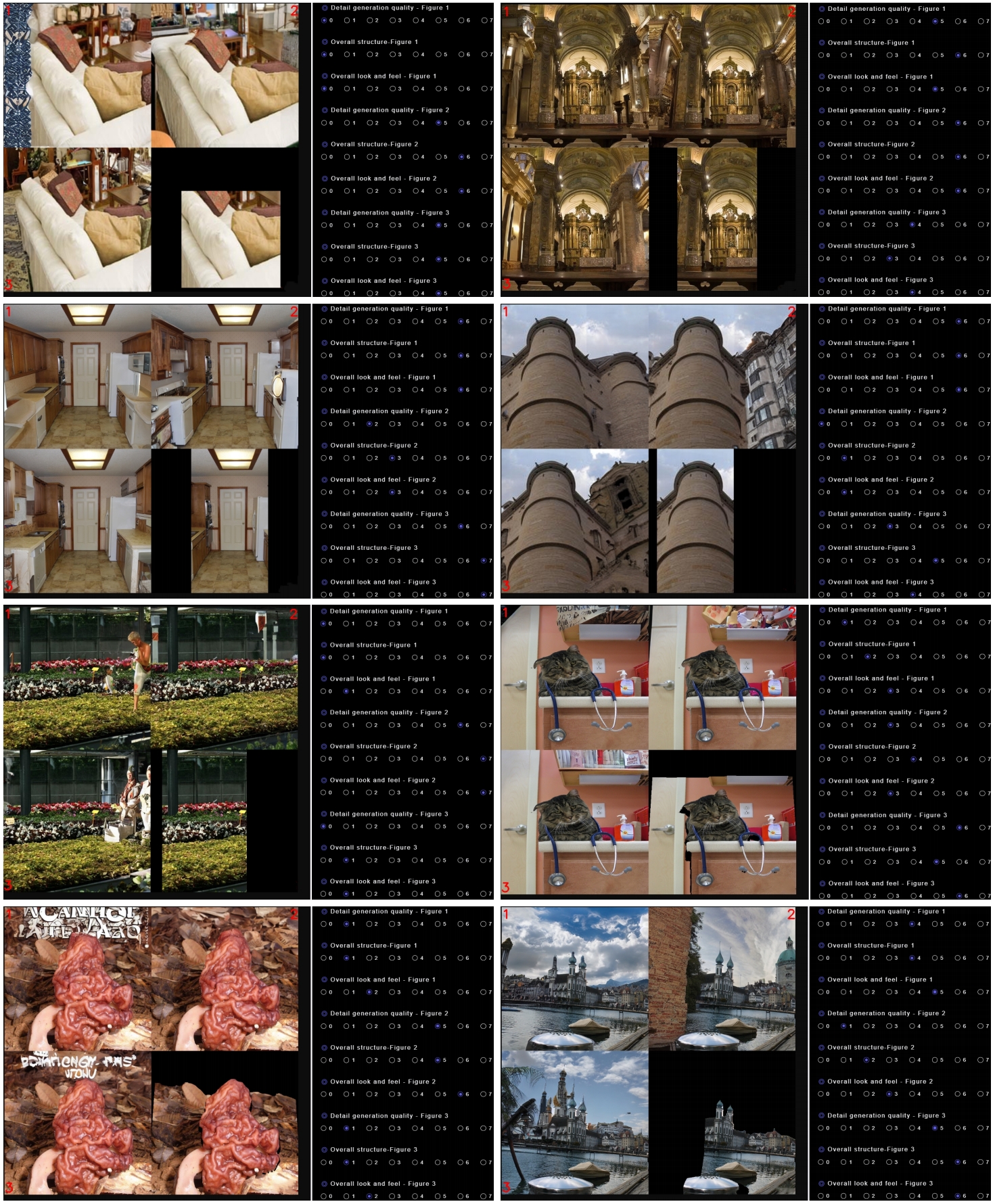}}
  \caption{\label{fig:labelplat}Labeling platform demonstration. In each group, from top left to bottom right, are reconstruction 1, 2, 3 and the prompt image.}
\end{figure}

\subsection{Labeling Platform}
The visual demonstration of labeling platform is shown in Fig.~\ref{fig:labelplat}. In the right-side, we aggregate three reconstruction results with the prompt images. The labeling is annotated in the left-side, where for each image, people need to label 3 scores by aforementioned 3 different criteria, ranging from 0 to 7.

\subsection{License}
The license of the proposed dataset is \textbf{CC BY-NC 4.0}, which allows creators to share their work with others while retaining certain rights but gives a restriction on commercial use.

\newpage
\section{More Application Visual Demonstrations}
To further verify the effectiveness of the proposed method, this section further visualizes more scenes for the application of the proposed method (as demonstrated in Sec.\ref{Sec:Applications}).

\subsection{Application of Image FOV Enlargement}
\begin{figure}[h]
    \centering
    \vspace{+0.2cm}
    \resizebox{\textwidth}{!}{\includegraphics{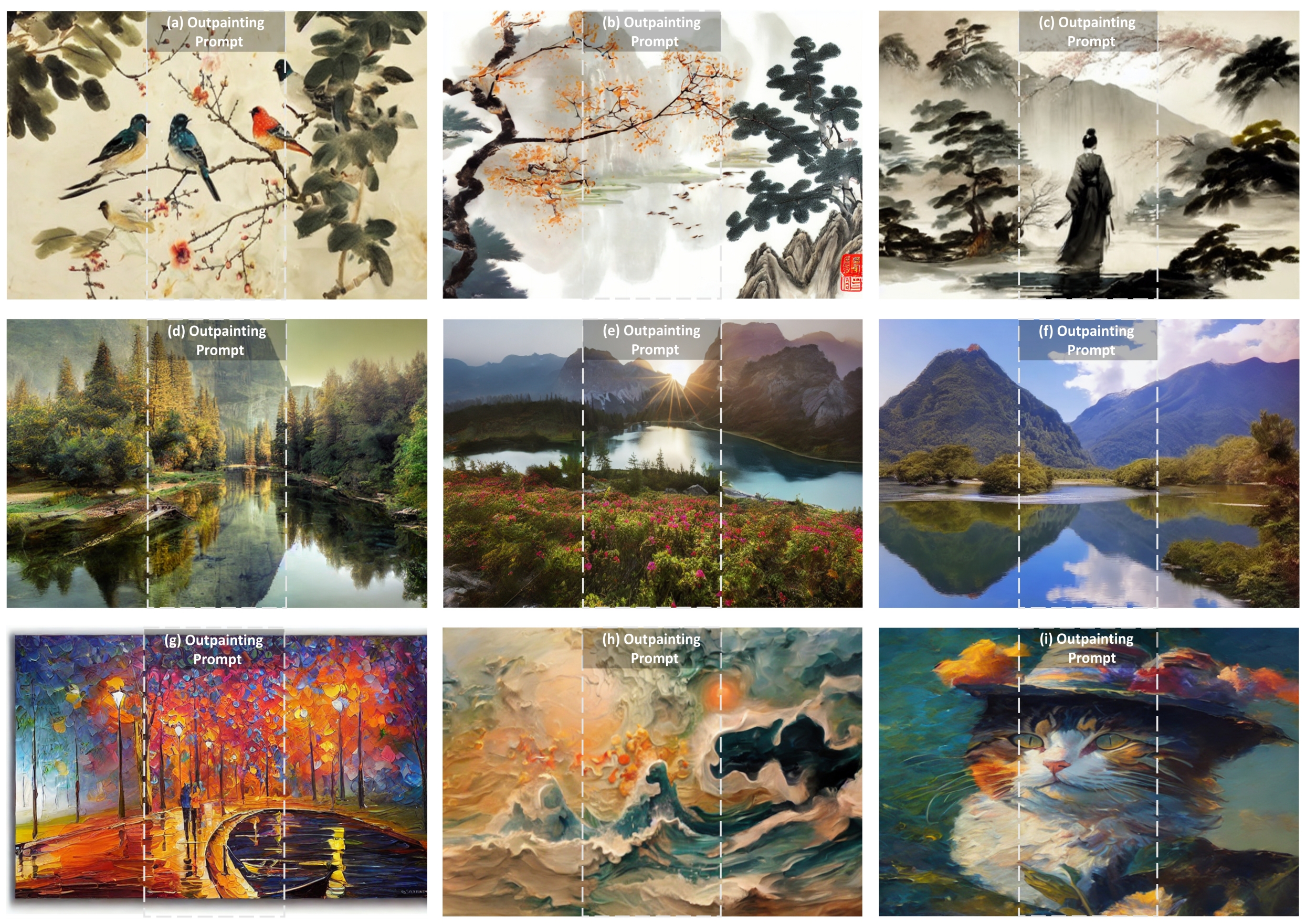}}
    \vspace{+0.2cm}
    \caption{ Application of FOV enlargement, where we visualize 9 scenes from (a) to (i) with corresponding prompt cropped image and enlarged result.} 
    \label{fig:exp3d_outpainting}
\end{figure}

For the task of Image FOV Enlargement, the results are shown in Fig.~\ref{fig:exp3d_outpainting}. We randomly select some picture from Internet and apply center crop on those pictures. We can see that no matter the style of prompt image, e.g., nature photography, oil painting or Chinese painting, the proposed method can consistently generate meaningful and visually pleasing results, which demonstrates the superority of the proposed method.

\vspace{+0.2cm}
\subsection{Application of Novel View Synthesis}
\begin{figure}[h]
    \centering
    \resizebox{\textwidth}{!}{\includegraphics{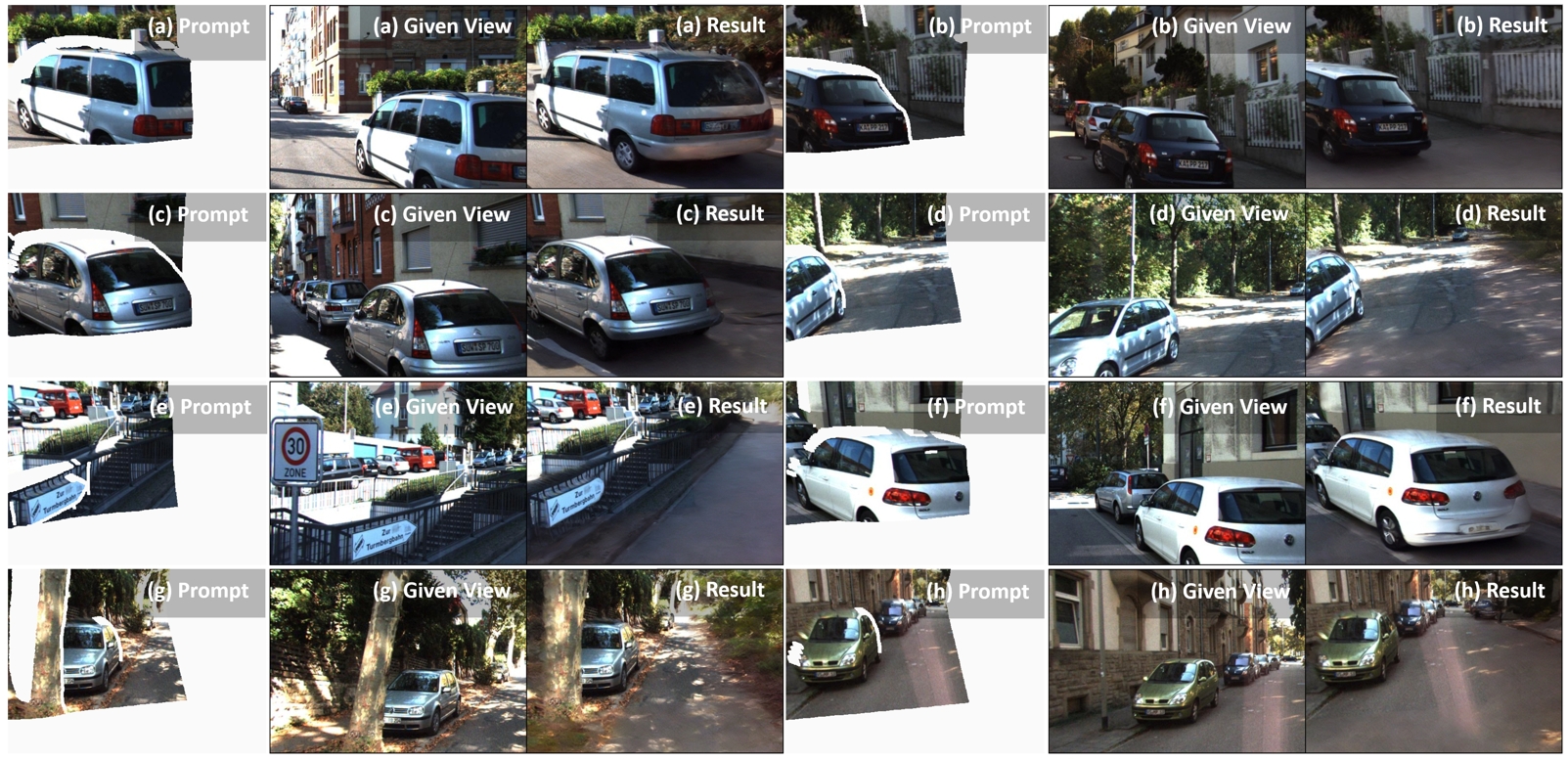}}
    \vspace{+0.2cm}
    \caption{ Application of novel view synthesis on KITTI dataset, where we visualize 8 scenes from (a) to (h) with corresponding prompt warped image, given view and reconstruction result.} 
    \vspace{+0.4cm}
    \label{fig:exp3d_KITTI}
\end{figure}

\begin{figure}[h]
    \centering
    \resizebox{\textwidth}{!}{\includegraphics{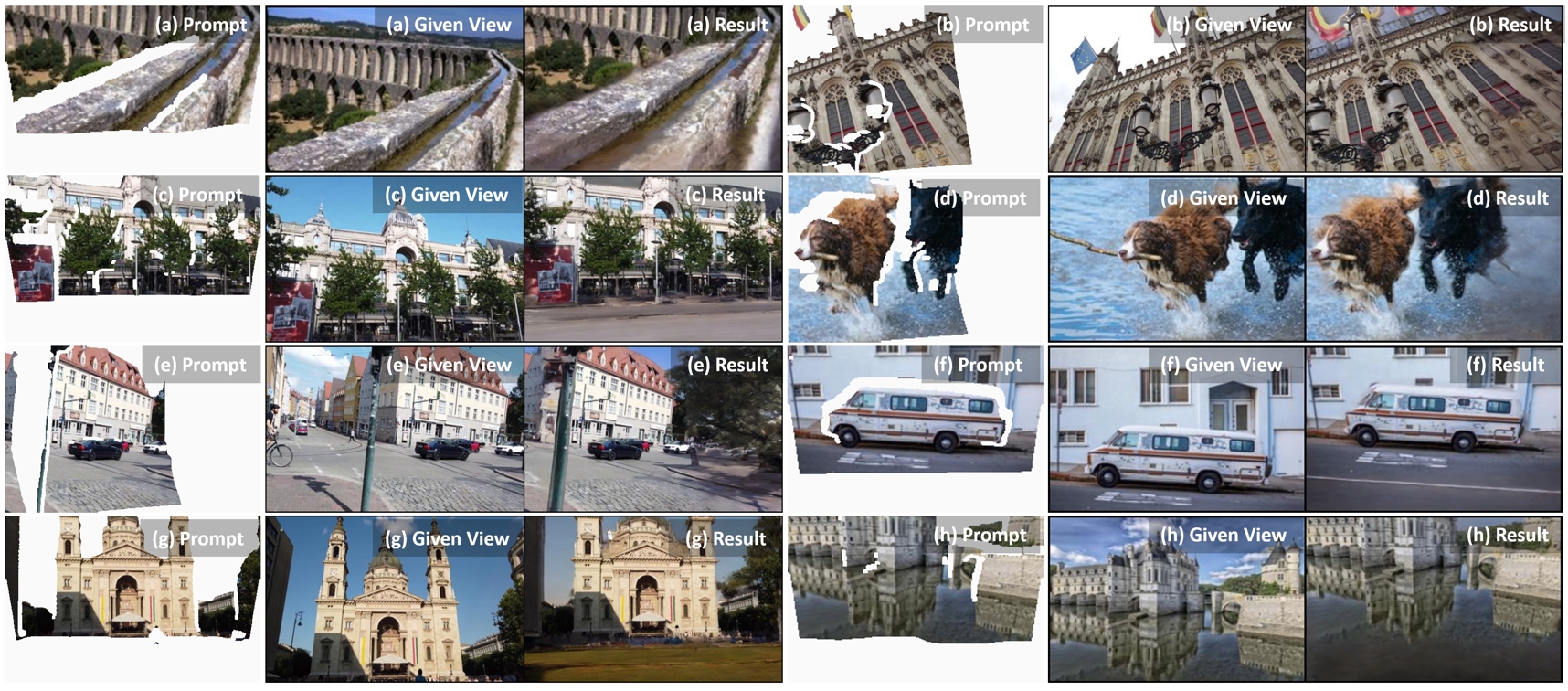}}
    \vspace{+0.2cm}
    \caption{ Application of novel view synthesis on DIV2K dataset, where we visualize 8 scenes from (a) to (h) with corresponding prompt warped image, given view and reconstruction result.}
    \vspace{+0.4cm} 
    \label{fig:exp3d_DIV}
\end{figure}

\vspace{+0.2cm}
We also visualize more novel view synthesis examples on the KITTI and DIV2K datasets in Figs.~\ref{fig:exp3d_KITTI} and \ref{fig:exp3d_DIV}. The warping-induced inpainting hole is much more irregular than the FOV Enlargement, which makes the task more challenging. However, the proposed method successfully filled up the missing region. The resulting high-quality reconstructions verify the superiority of the proposed method.

\newpage
\section{Reward Scoring Statistics \& Reward Normalization Configuration}

\subsection{Score Distribution across Different Categories}
\begin{figure}[htbp]
  \centering
    \includegraphics[height=10cm, width=\linewidth]{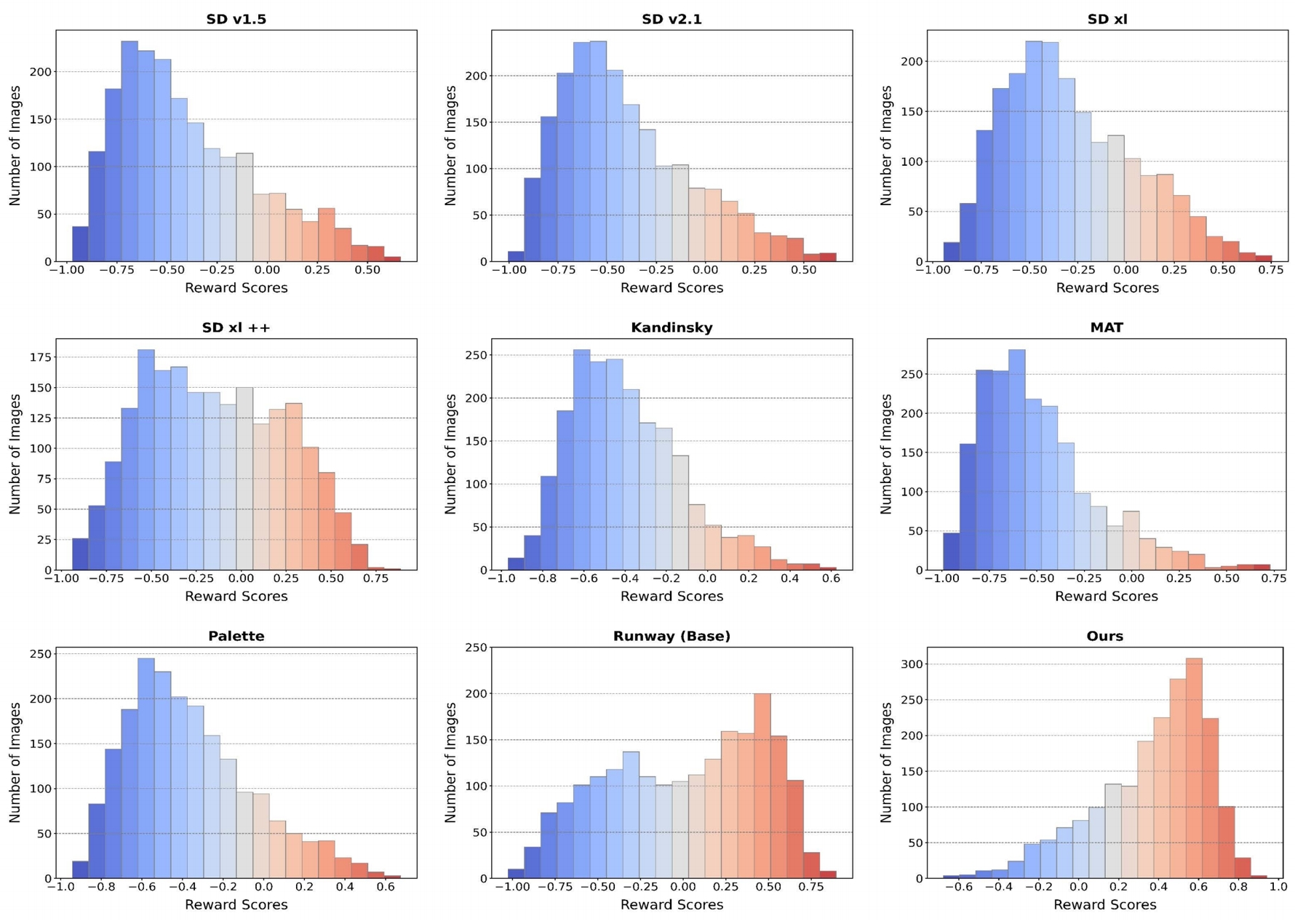}
  \caption{\label{fig:supp_rewards} Detailed score statistics of the proposed dataset.}
\end{figure}

We count the ratio of prompt with different semantic class in Fig.~\ref{fig:supp_rewards}. Most methods, including SOTA diffusion-based methods, e.g., stable diffusion or Runway, have a large ratio of samples with negative rewards. However, the proposed method achieves the most positive rewards, as shown in the last figure, which validates the necessity and effectiveness of applying such an alignment task to the diffusion in-painting model.

\subsection{Normalization Factors} To facilitate the training of reward model, we normalize the score with $\frac{s-mean}{var}$, where $s$ is raw score data. To remove the data bias for training the reward model, the normalization factors are calculated within different datasets and inpainting patterns. We show the resulting factors in Table ~\ref{tab:data_normalize}. The relatively balanced factors indicate that our labels do not have large biases on different datasets or inpainting pattern.

\begin{table}[htbp]
    \centering
    \renewcommand{\arraystretch}{1.3}
    \setlength{\tabcolsep}{1pt} 
    \normalsize
    \begin{tabular}{c|cc|cc|cc|cc}
    \toprule
        \multirow{2}{*}{\textbf{Dataset}}  &  \multicolumn{2}{c|}{ADE20K}&  \multicolumn{2}{c|}{KITTI}&  \multicolumn{2}{c|}{ImageNet}&  \multicolumn{2}{c}{Div2K} \\
            & Warping & Outpainting & Warping & Outpainting & Warping & Outpainting & Warping & Outpainting              \\ \midrule
        \textbf{Mean}    & 3.46 & 3.12 & 3.02 & 2.87 & 2.85 & 2.50 & 2.99 & 2.34  \\
        \textbf{Variance} & 2.77 & 4.42 & 3.04 & 2.69 & 3.03 & 3.08 & 3.26 & 3.60 \\
        \bottomrule
    \end{tabular}
    \vspace{0.5cm}
    \caption{The normalization factor of score to facilitate for reward model training.}
    \label{tab:data_normalize}
\end{table}

\section{More Visualizations}
\subsection{More Visualizations of Comparisons}
\begin{figure}[htbp]
  \centering
  \includegraphics[height = 21cm, width=\linewidth]{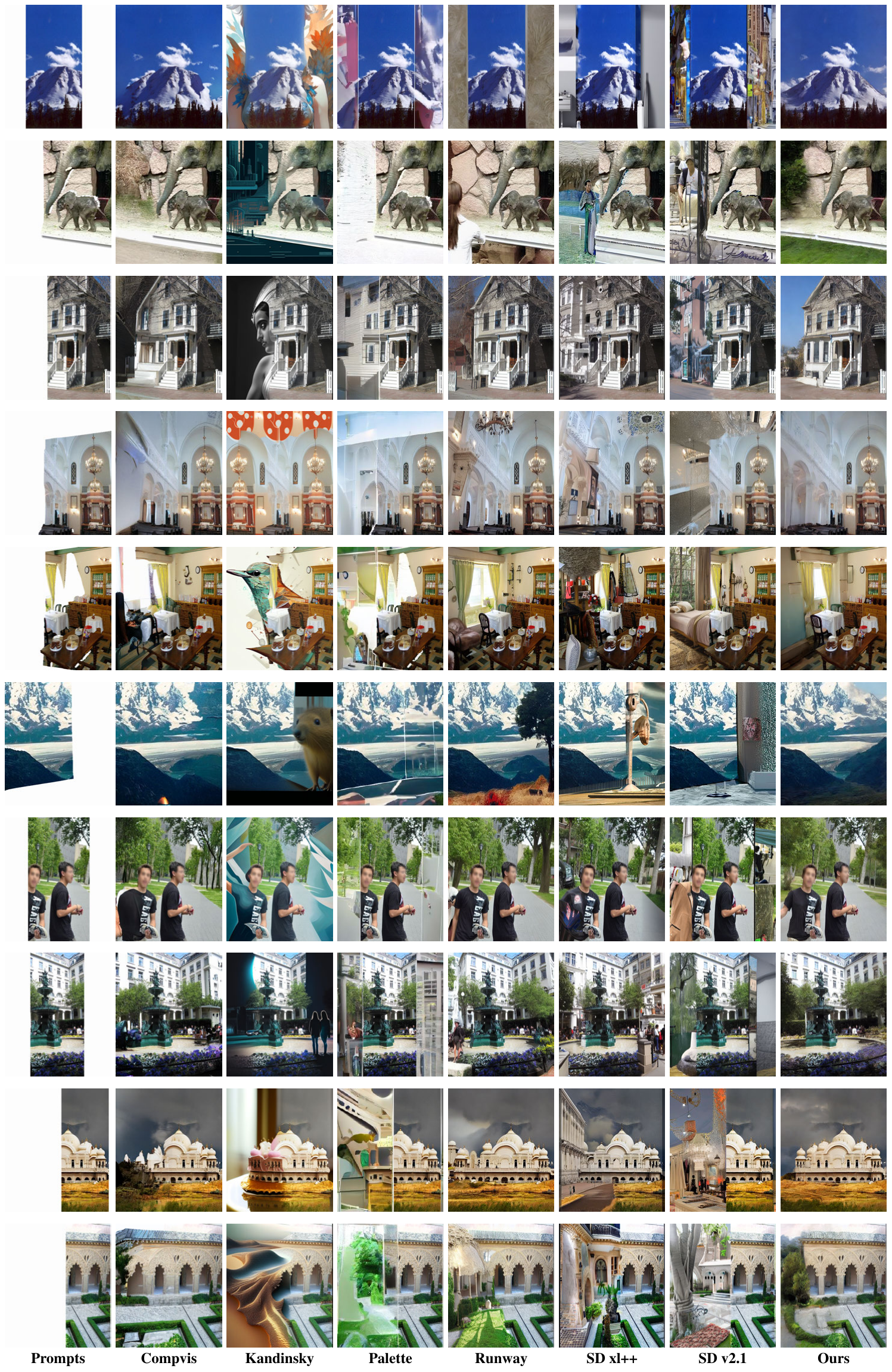}
  \caption{\label{fig:supp_comparisons_2}Qualitative comparison between the proposed method and other SOTA methods.}
\end{figure}

\begin{figure}[htbp]
  \centering
  \includegraphics[height = 21cm, width=\linewidth]{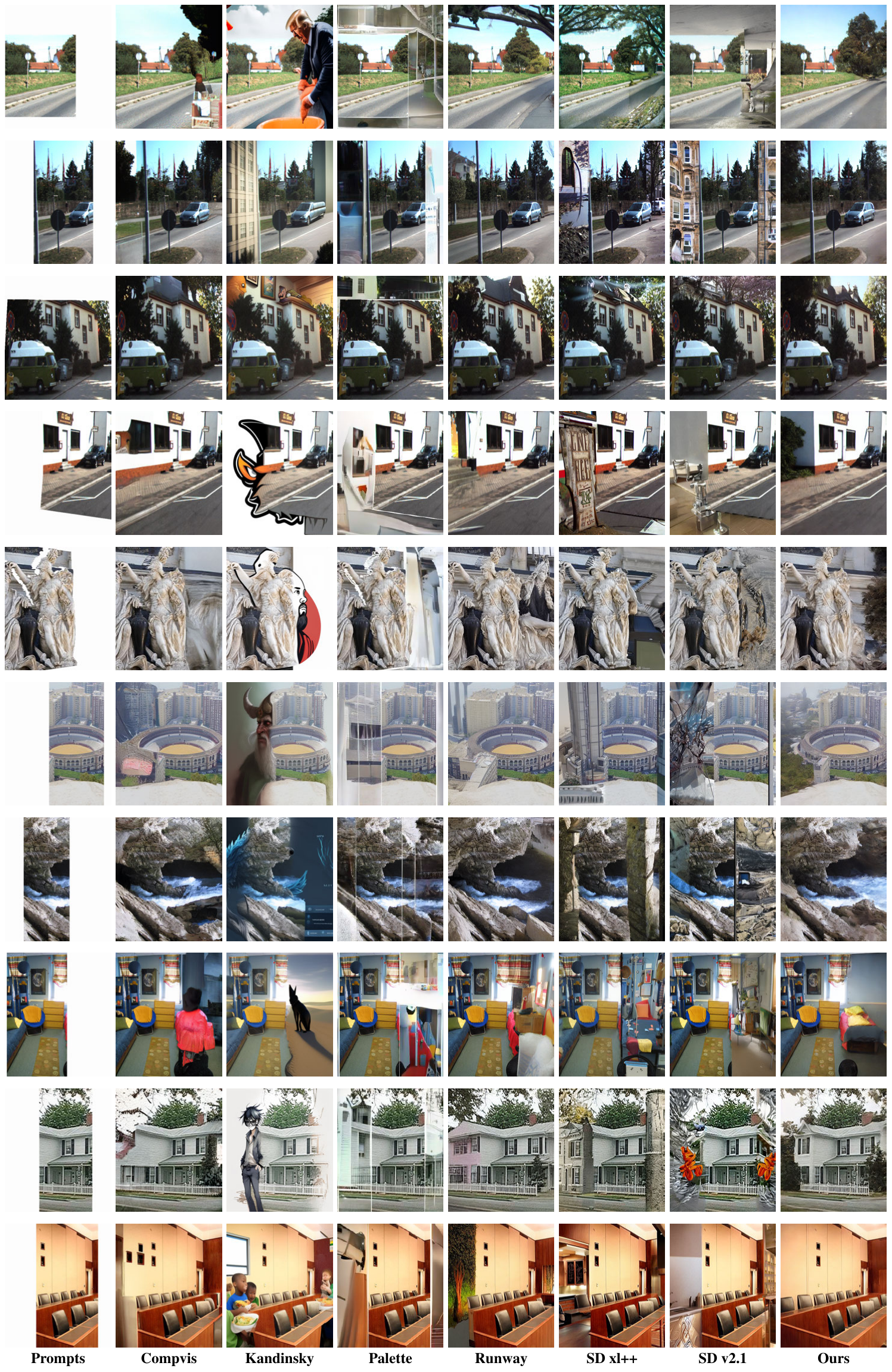}
  \caption{\label{fig:supp_comparisons_1}Qualitative comparison between the proposed method and other SOTA methods.}
\end{figure}

In this section, we visually compare the reconstruction results of different methods. As shown in Figs.~\ref{fig:supp_comparisons_2} and \ref{fig:supp_comparisons_1}. The comparison of different methods in Figs.~\ref{fig:supp_comparisons_2} and \ref{fig:supp_comparisons_1} indicate that the proposed method could generate high-quality reconstruction compared with other SOTA methods.
Notably, the results from our approach show improved clarity and detail retention, which are essential for practical applications. 

\subsection{More Visualizations of Multiple Sampling}

\begin{figure}[htbp]
  \centering
  \includegraphics[height = 16cm, width=\linewidth]{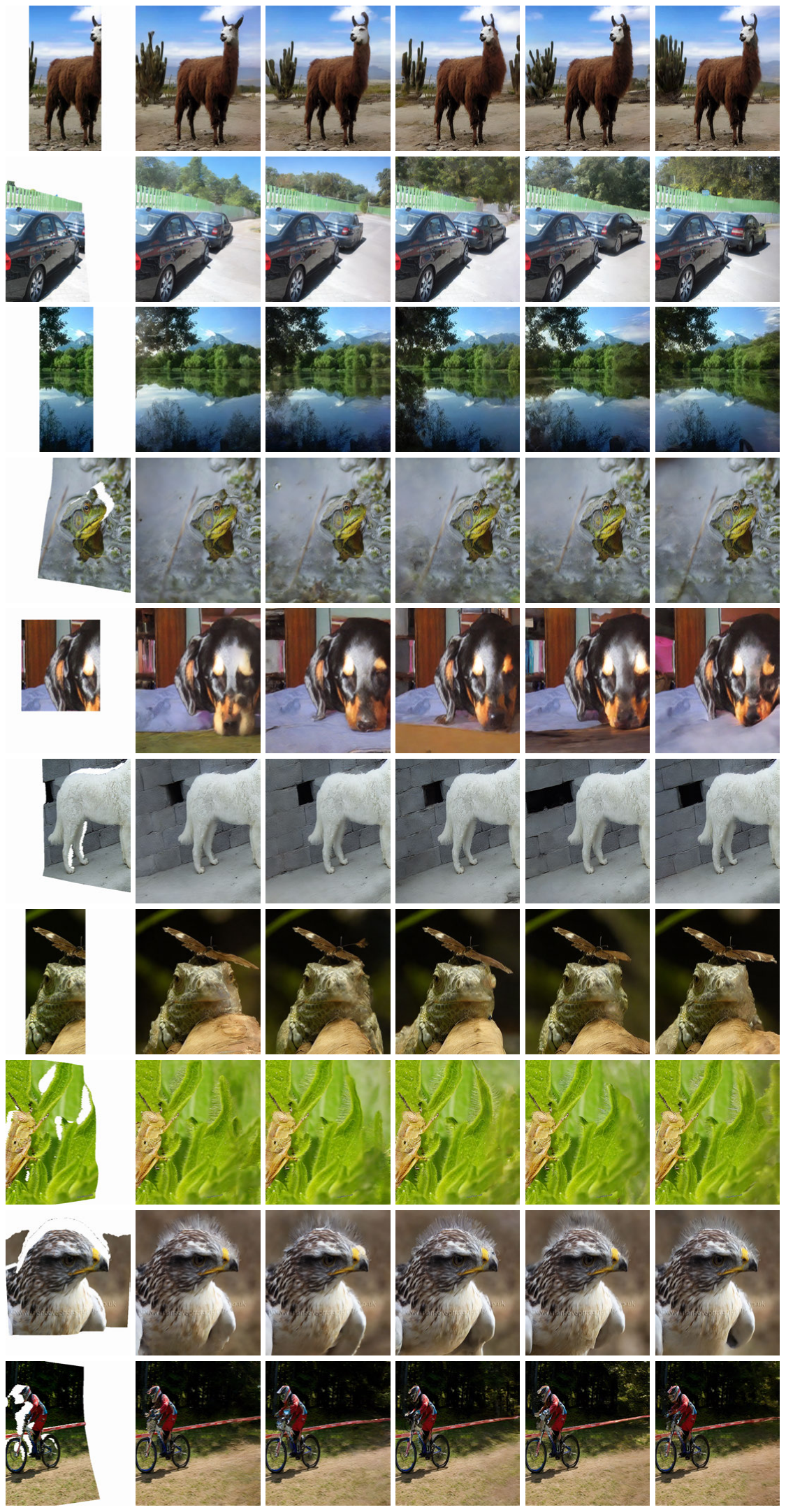}
  \caption{\label{fig:supp_sampling}Qualitative comparison of different sampling times of the proposed method.}
\end{figure}

To verify the robustness and consistency of the inpainting model after alignment with our method. We run 5 times with different random seeds on the same prompt to derive different reconstruction results. The results shown in Fig.~\ref{fig:supp_sampling} demonstrate the stability of the proposed method. We can see the proposed method generate high-quality and visually pleasing results under different conditions.

\section{More experimental results}
\setcounter{figure}{0}   
\setcounter{table}{0}  
\subsection{Comparisons with Additional inpainting Methods}
We have experimentally compared with some inpainting Methods, i.e., PowerPaint~\cite{powerpaint}, BrushNet~\cite{brushnet} and Hdpaint~\cite{hdpaint} . For all these methods, we assessed performance using their publicly released models. As shown in the Table~\ref{tab_subject} and Table~\ref{tab_winrate}, our method significantly outperforms all the compared methods, particularly in perceptual metrics such as T2I and our Reward score. Besides, BrushNet employs roughly twice the number of parameters as our model, and HdPaint requires 15 times more computational time (experimented on NVIDIA GeForce RTX A6000 GPUs), making it impractical for real-world applications.

\begin{table}[htb]
\centering
\renewcommand{\arraystretch}{1.3}
\setlength{\tabcolsep}{2pt} 
\caption{\label{tab_subject}Comparison across metrics: higher values are better for all metrics except "\# Param".}
\vspace{0.2cm}
\begin{tabular}{c|c|c|c|c|c|c|c}
\toprule
Metrics & T2I & BLIP & CLIP & CA(Incep.)  &  Reward & \# Param(M) & Infer. Time(s) \\
\midrule
PowerPaint(v-1) & -4.44 & 0.46 & 0.21 & 0.42  & -0.057 & 819.72 & \textbf{5} \\
PowerPaint(v-BrushNet) & -3.84 & 0.46 & 0.20 & 0.42  & -0.036 & 1409.88 & 16\\
BrushNet & 1.26 & 0.46 & 0.22 & 0.43  & 0.137 & 1409.86 & 16\\
HdPaint & -4.57 & 0.47 & 0.21 & 0.44  & -0.059 & \textbf{451.47} & 60\\
\textbf{PrefPaint(Ours)} & \textbf{11.60} & \textbf{0.49} & \textbf{0.23} & \textbf{0.45} & \textbf{0.374} & 819.72 & \textbf{5}\\
\bottomrule
\end{tabular}
\end{table}

\begin{table}[h]
\centering
\renewcommand{\arraystretch}{1.3}
\setlength{\tabcolsep}{2pt} 
\caption{\label{tab_winrate} Quantitative comparisons of different methods. ``$S$'' is the number of sampling times. For the calculation of WinRate, we first derive the best sample of the compared method among $S$ sampling times. }
\begin{tabular}{c|c|c|c}
\toprule
WinRate$\uparrow$ (v.s. Runway) & S=1 & S=2 & S=3 \\
\midrule
PowerPaint(v-1) & 27.06\% & 39.92\% & 47.38\%  \\
PowerPaint(v-BrushNet) & 29.86\% & 43.12\% & 52.01\% \\
BrushNet & 49.49\% & 62.83\% & 69.22\%  \\
HdPaint(ds8-inp) & 33.37\% & 43.41\% & 49.03\%  \\
\textbf{PrefPaint(Ours)} & \textbf{71.27\%} & \textbf{85.88\%} & \textbf{93.50\%} \\
\bottomrule
\end{tabular}
\end{table}

\subsection{Comparisons with other reinforcement learning methods.}

we have experimentally compared our method with some other reinforcement learning methods in Table~\ref{tab_dpo}. Specifically, we simply summarize the implementation of each method. Human Preference Score~\cite{wu2023human} learns a negative prompt to map the diffusion process to low-quality samples. Then, in the inference process, the negative sample is utilized in the classifier-free guidance~(CFG) to push the generation trajectory away from low-quality samples. ImageReward~\cite{xu2024imagereward} trains a reward model and then applies the reward model as a loss metric to end-to-end optimize the diffusion model accompanied by a reconstruction loss. We also conduct the ablation study on reward training strategy in Table 3 in our paper. Our method employs a regression-driven training strategy, while ImageReward a classification-drive strategy. DPOK~\cite{fan2024reinforcement} simultaneously optimizes the whole trajectory of a reverse diffusion process and utilizes the KL divergence to panel the regularization, avoiding a large distribution shift.
D3PO~\cite{yang2024using} adopts the RL strategy from direct performance optimization (DPOK), directly optimizes the model on the reward labeling data to minimize the probability of low-quality samples and increase the probability of high-quality samples.

\begin{table}[h]
\centering
\renewcommand{\arraystretch}{1.3}
\setlength{\tabcolsep}{2pt} 
\caption{\label{tab_dpo} Quantity evaluations of different reinforcement learning methods.}
\vspace{0.2cm}
\begin{tabular}{c|c|c|c|c|c|c}
\toprule
Methods & WinRate & T2I & Reward  & CLIP & BLIP & CA\\
\midrule
Human Preference Score & 58.03\% & -16.67 & 0.26 & 0.20& 0.47 & 0.40  \\
ImageReward & 65.10\% & 13.12 & 0.29 & 0.22& 0.48& 0.44\\
DPOK (KL weight=0.1) & 64.59\% & 11.43 & 0.32 & 0.21& 0.48& 0.43 \\
DPOK (KL weight=1.0)& 62.67\% & 9.36 & 0.30 &0.21 & 0.48& 0.43\\
D3PO & 59.74\% & -19.20 & 0.26 &0.21 & 0.46 & 0.41\\
\textbf{PrefPaint} & \textbf{71.27\%} & \textbf{21.53} & \textbf{0.37} & \textbf{0.23} & \textbf{0.49} & \textbf{0.45} \\
\bottomrule
\end{tabular}
\end{table}

\subsection{Compared with SD xl ++}
Currently, their is also a kind of diffusion-inpainting network like SD xl ++~\cite{sdxlInpt}, which requires a complete image to redraw the mask region and can generate a visually pleasing result. However, their performance would be dramatically reduced without the complete image. To validate the characteristics of SD xl ++, we carry out experiments to examine its performance. 

\begin{table*}[htbp]
\renewcommand{\arraystretch}{1.2}
\setlength{\tabcolsep}{6.5pt} 
\centering
\small
\caption{\label{SUPPTable:comparisons}Alignment evaluation across various generative methods, where $\star$ represents the small model~(non SD-based); ``$S$'' refers to the number of sampling times. For comparing the WinRate, the best inpainting results are selected through multiple sampling and compared with the \textit{Runway} Inpainting model with only 1 sampling. ``$\uparrow$ (resp. $\downarrow$)" means the larger (resp. smaller), the better. The WinRate is calculated against Runway~\cite{runway}.} 
\begin{tabular}{c|cc|cc} 
\toprule
\multirow{1}{*}{Prompt Methods} & \multicolumn{2}{c|}{Outpainting Prompts} & \multicolumn{2}{c}{Warping Prompts} \\
 \cmidrule{2-5} 
\multirow{2}{*}{Metrics}  & \multicolumn{1}{c}{WinRate} & \multicolumn{1}{c|}{Rewards}& \multicolumn{1}{c}{WinRate} & \multicolumn{1}{c}{Rewards} \\ 
&  (\%)~$\uparrow$ & Mean$\uparrow$  &  (\%)~$\uparrow$ &Mean$\uparrow$   \\ \midrule
SD v1.5 w/ xl ++ & 21.15  & -0.13  & 18.66 & -0.18  \\ 
SD v2.1 w/ xl ++& 26.07  & -0.07  & 22.26 & -0.12  \\  
Kandinsky w/ xl ++ & 17.09  & -0.29   & 12.04  &  -0.37  \\  
MAT $\star$ w/ xl ++ & 19.12  & -0.34   & 8.58 & -0.56  \\  
Palette $\star$ w/ xl ++ & 27.99  & -0.07  & 25.82 & -0.09  \\  
Runway w/ xl ++ & 56.94 & +0.03 & 45.71 & +0.02 \\ \midrule
\textbf{Ours} & 70.16 & +0.38 & 72.38 & +0.36 \\
\bottomrule
\end{tabular}
\vspace{-0.3cm}
\end{table*}

\section{Details \& Reward}
\vspace{-0.2cm}
\setcounter{figure}{0}   
\setcounter{table}{0}   

\subsection{Training Curve}
We present training curves of various methods for comparison. The curves for '1.4BaseLine' and '1.4Boundary' (Ours) demonstrate our modification's acceleration, reaching 0.35 rewards first. Our approach converges with the fewest training iterations.

\begin{figure}[htb]
  \centering
  \includegraphics[width=0.9\linewidth]{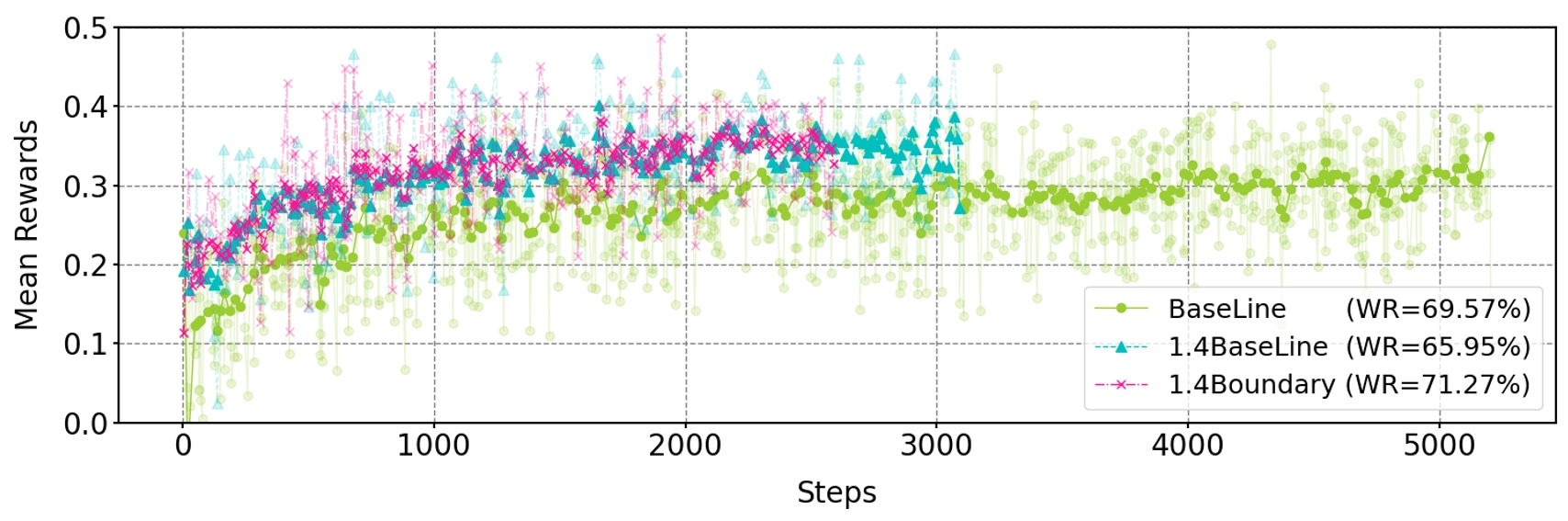}
  \vspace{-0.3cm}
  \caption{\label{fig:training_curbe} Training curves of different experimental setups.}
  \vspace{-0.3cm}
\end{figure}

\subsection{Details between Eq. 4-5} According to the~\cite{abbasi2011improved}, we derive a similar process for the linear reward regression on feature space, where $\mathbf{Z}$ denotes the latent of an image.
\begin{align}
\hat{\boldsymbol{\psi}}=& (\mathbf{Z}^T\mathbf{Z} + \lambda \mathbf{I})^{-1}\mathbf{Z}^T\mathbf{Y}\\
=&(\mathbf{Z}^T\mathbf{Z} + \lambda \mathbf{I})^{-1}\mathbf{Z}^T(\mathbf{Z}\boldsymbol{\psi}_* + \boldsymbol{\zeta}) \\
=&(\mathbf{Z}^T\mathbf{Z} + \lambda \mathbf{I})^{-1}[(\mathbf{Z}^T\mathbf{Z} + \lambda \mathbf{I} - \lambda \mathbf{I})\boldsymbol{\psi}_* + \mathbf{Z}^T\boldsymbol{\zeta}] \\
=& \boldsymbol{\psi}_* - \lambda(\mathbf{Z}^T\mathbf{Z} + \lambda \mathbf{I})^{-1} \boldsymbol{\psi}_* + (\mathbf{Z}^T\mathbf{Z} + \lambda \mathbf{I})^{-1} \boldsymbol{\zeta}
\end{align}
\begin{align}
    \hat{\boldsymbol{\psi}} - \boldsymbol{\psi}_* =&(\mathbf{Z}^T\mathbf{Z} + \lambda \mathbf{I})^{-1} \mathbf{Z}^T \boldsymbol{\zeta} - \lambda (\mathbf{Z}^T\mathbf{Z} + \lambda \mathbf{I})^{-1} \boldsymbol{\psi}_*,
\end{align}

\end{document}